\PassOptionsToPackage{dvipsnames,table}{xcolor}

\documentclass[runningheads,dvipsnames,table]{llncs}

\usepackage{eccv}

\usepackage{eccvabbrv}

\usepackage{graphicx}
\usepackage{booktabs}
\usepackage{microtype}
\usepackage{xspace}

\usepackage{tabularx}
\newcolumntype{Y}{>{\centering\arraybackslash}X}
\usepackage{array,multirow}
\usepackage{amsmath}
\usepackage[bb=dsserif]{mathalpha}
\usepackage{bm}
\usepackage{enumitem}
\usepackage{courier}
\usepackage{mathtools}
\usepackage{lipsum}
\usepackage{wrapfig}

\usepackage{tikz}
\tikzset{fontscale/.style = {font=\relsize{#1}}
    }
    
\usepackage[subtle]{savetrees}
\usepackage{pifont}

\newcommand{\ours}{GalLoP\xspace}

\newcommand{\gallopemoji}{\includegraphics[height=0.6em]{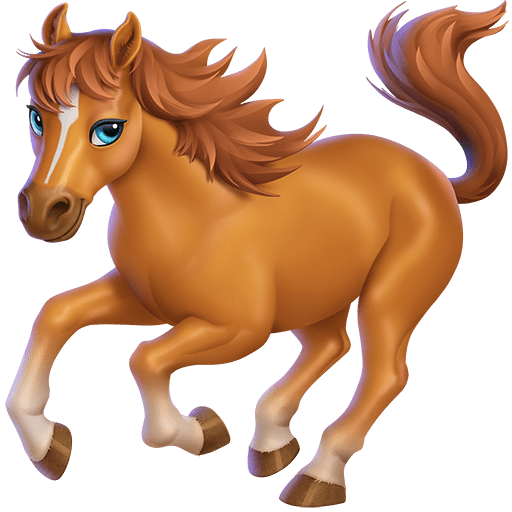}\xspace}

\definecolor{mycornflowerblue}{rgb}{1, 0.95, 0.9}

\usepackage[accsupp]{axessibility}  
 \usepackage{colortbl}

\usepackage[pagebackref,breaklinks,colorlinks]{hyperref}

\usepackage{orcidlink}

\begin{document}

\title{\gallopemoji \ours: Learning Global and Local Prompts \\ for Vision-Language Models} 

\titlerunning{\ours: Learning Global and Local Prompts for VLMs}

\author{Marc Lafon$^\star$\inst{1}\orcidlink{0009-0002-0688-177X}, Elias Ramzi$^\star$\inst{1}\orcidlink{0000-0002-0131-2458},\\
Clément Rambour\inst{1}\orcidlink{0000-0002-9899-3201}, 
Nicolas Audebert\inst{1,2}\orcidlink{0000-0001-6486-3102},
\and Nicolas Thome\inst{3}\orcidlink{0000-0003-4871-3045}}

\authorrunning{M.~Lafon et al.}

\institute{
Conservatoire national des arts et métiers, CEDRIC, F-75141 Paris, France
\and
Univ. Gustave Eiffel, ENSG, IGN, LASTIG, F-94160 Saint-Mandé, France
\and
Sorbonne Université, CNRS, ISIR, F-75005 Paris, France\\
\email{\{marc.lafon, elias.ramzi\}@lecnam.net}\\
}

\maketitle

\newcommand\blfootnote[1]{%
  \begingroup
  \renewcommand\thefootnote{}\footnote{#1}%
  \addtocounter{footnote}{-1}%
  \endgroup
}

\blfootnote{$^\star$ Equal contribution.}

\begin{abstract}
    
    Prompt learning has been widely adopted to efficiently adapt vision-language models (VLMs), \eg CLIP, for few-shot image classification. Despite their success, most prompt learning methods trade-off between classification accuracy and robustness, \eg in domain generalization or out-of-distribution (OOD) detection. In this work, we introduce \textbf{G}lob\textbf{al}-\textbf{Lo}cal \textbf{P}rompts (\ours), a new prompt learning method that learns multiple diverse prompts leveraging both global and local visual features. The training of the local prompts relies on local features with an enhanced vision-text alignment. To focus only on pertinent features, this local alignment is coupled with a sparsity strategy in the selection of the local features. We enforce diversity on the set of prompts using a new ``prompt dropout'' technique and a multiscale strategy on the local prompts. \ours outperforms previous prompt learning methods on accuracy on eleven datasets in different few shots settings and with various backbones. Furthermore, \ours shows strong robustness performances in both domain generalization and OOD detection, even outperforming dedicated OOD detection methods. Code and instructions to reproduce our results: \url{https://github.com/MarcLafon/gallop}.
  
  \keywords{Vision-language models \and Few shot classification \and Prompt learning \and Local and global prompts \and Robustness \and OOD detection} 
\end{abstract}

\vspace{2em}
\section{Introduction}\label{sec:intro}

Vision-Language Models (VLMs), \eg CLIP~\cite{Radford21} or ALIGN~\cite{jia2021scaling}, have shown impressive performances for zero-shot image classification. Prompt learning \cite{zhou2022learning, zhou2022conditional, Chen22, khattak2023maple, khattak2023self, parisot2023learning, lu2022prompt} has been among the leading approaches to efficiently adapt VLMs to a specific downstream dataset. These methods train a learnable context in the form of \textit{soft prompts} to optimize the text/image alignment.
~Prompt learning methods benefit from the strong generalization capability of VLMs'
~textual encoder and are effective even when only a few labeled examples are available.

\begin{figure}[t]

\begin{tabular}{ccc}
\includegraphics[width=0.33\linewidth, height=0.33\linewidth]{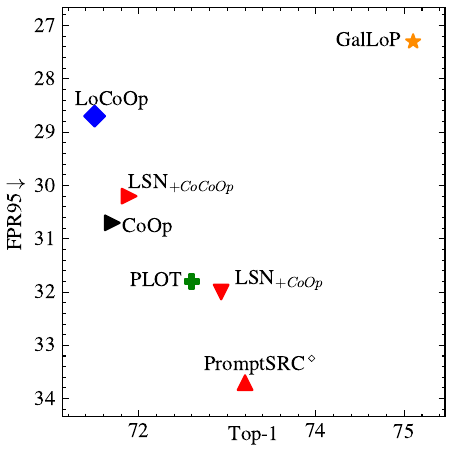} &
\includegraphics[width=0.33\linewidth, height=0.33\linewidth]{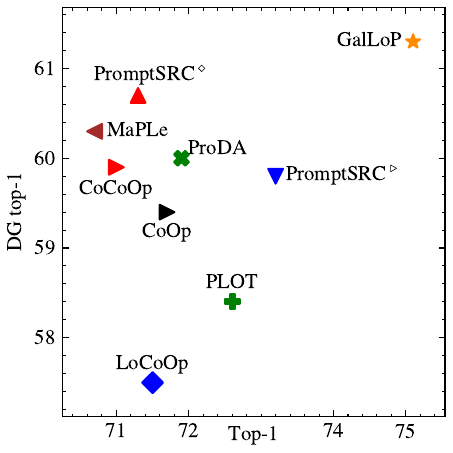} &
\includegraphics[width=0.33\linewidth, height=0.33\linewidth]{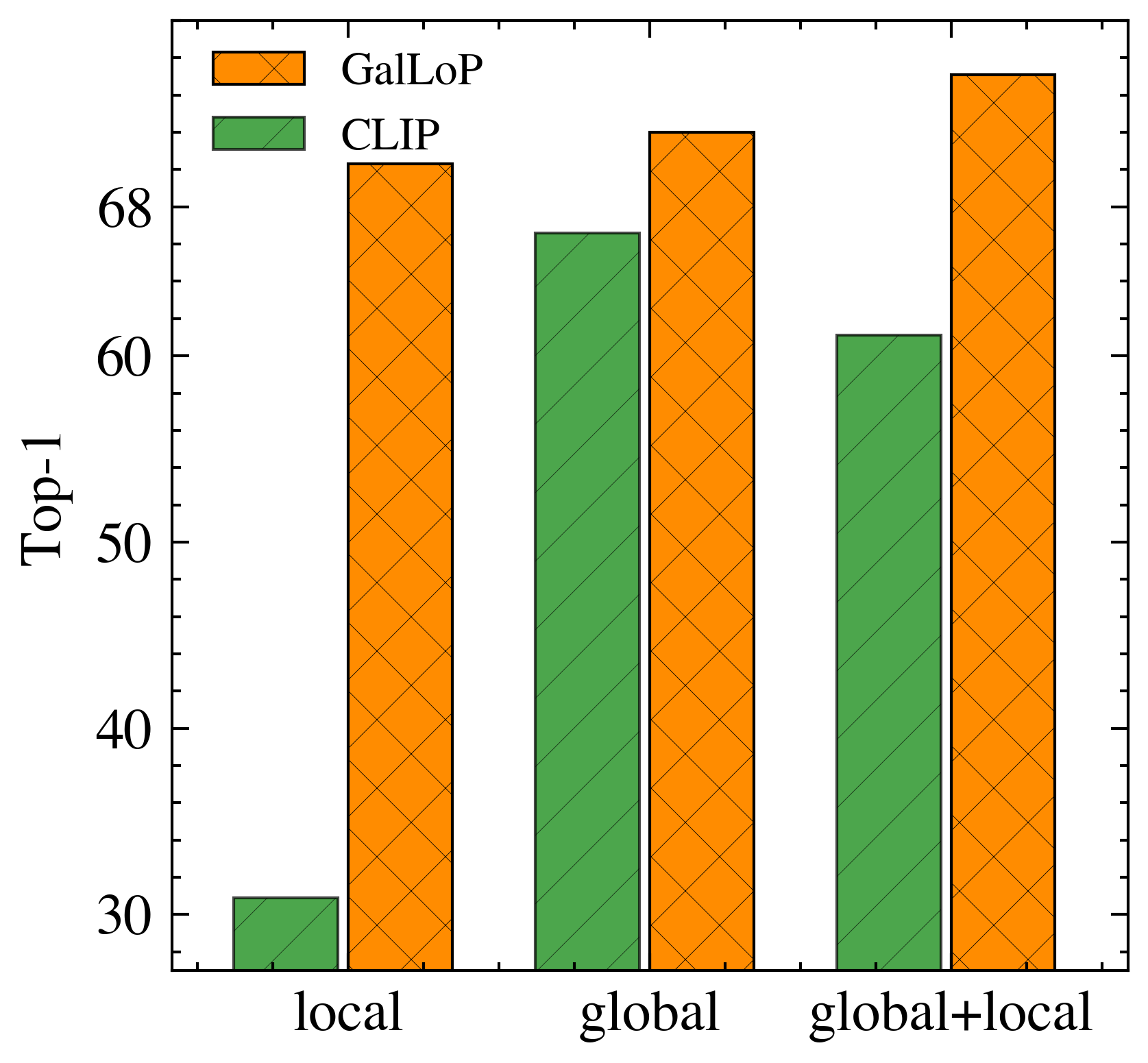} \\
(a) OOD detection & (b)  Domain Generalization & (c) Global-local  
\end{tabular}

\caption{Our \ours method demonstrates excellent performances in accuracy plus robustness, \ie out-of-distribution detection (a) and domain generalization (b), while state-of-the-art prompt learning methods compromise between these aspects. Additionally, unlike recent methods utilizing ineffective local zero-shot CLIP features, \ours learns discriminative local prompts precisely aligned with sparse image regions at various scales, facilitating the discriminability between classes. \ours integrates both global and local prompts, with their diversity explicitly enforced during few-shot learning, which significantly enhances the performance of their combination (c). 
}
\label{fig:figure_intro}
\end{figure}

Despite their success, we observe that these methods trade off between classification accuracy and robustness.
~This is illustrated on~\cref{fig:figure_intro}(a), where methods exhibiting the best accuracy sacrifice out-of-distribution (OOD) detection performances, \eg PromptSRC~\cite{khattak2023self}, while those excelling in OOD detection often have poor accuracy results, \eg LoCoOp~\cite{Miyai24}. A similar observation is done in domain generalization (DG), see~\cref{fig:figure_intro}(b): PromptSRC \cite{khattak2023self} presents two different versions, one optimized for accuracy (PromptSRC$^\triangleright$) and the other for domain generalization (PromptSRC$^\diamond$), highlighting the intrinsic conflict between both criteria.

To boost classification accuracy, prompt learning can involve learning multiple prompts \cite{agnolucci2023eco} to emulate ``prompt ensembling'', \eg prompts specialized for specific classes ~\cite{parisot2023learning, wang2022learning} or Transformer's layers \cite{khattak2023maple, khattak2023self}, or casting multiple prompts learning within a probabilistic framework~\cite{lu2022prompt}.
~The key challenge in prompt ensembling lies in learning diverse prompts to optimize the combination. However, since these approaches only operate on global visual representations, they cannot utilize diverse prompts aligned with specific image regions to maximize their diversity.

Recently, attempts have been made to use local image representations in prompt learning, \eg LoCoOp \cite{Miyai24} or PLOT \cite{Chen22}.
Although these approaches are promising, their performances in accuracy/robustness are suboptimal compared to state-of-the-art results, see~\cref{fig:figure_intro}(a),(b). Their limited performances stem from two main factors: i) they use ``dense'' (\ie all) local features from CLIP, which includes irrelevant or noisy regions for a given concept, and ii) these local features are not as well aligned with the text due to CLIP's pre-training with the global representation. In consequence, the performance of prompts trained with those local features is much lower than their global counterpart, and this degradation affects performances when combined with global, as illustrated in~\cref{fig:figure_intro}(c).

In this paper, we introduce \textbf{G}lob\textbf{al}-\textbf{Lo}cal \textbf{P}rompts (\ours), a new method to learn a diverse set of prompts by leveraging both global and local visual representations. \ours learns sparse discriminative local features, \ie text prompts are aligned to a sparse subset of regions at multiple scales. This enables fine-grained and accurate text-to-image matching, making \ours local prompts highly competitive. Moreover, we train \ours with diverse global and local prompts, ~unlocking the complementarity between both sets and significantly improving their combination, as shown in~\cref{fig:figure_intro}(c). 

To achieve this, \ours relies on two main methodological contributions:
\begin{itemize}
    \item \textbf{Effective local prompts learning.} In \ours, we propose to align local prompts with sparse subsets of $k$ image regions, enabling text-to-image matching that captures fine-grained semantics. To adapt visual representations to the downstream dataset, we ~refine the textual alignment of visual local features ~by employing a simple linear projection amenable to few-shots learning. 
    \item \textbf{Enforcing ensemble diversity.} We learn both global prompts aligned with the whole image and local spatially-localized prompts, and enforce diversity between them to improve their combination. We induce diversity through randomization using a new ``prompt dropout'' strategy, which enhances generalization when learning multiple prompts. Additionally, we employ a multiscale strategy to align local prompts with image regions of varying sizes, capturing different visual aspects of a concept's semantics.
\end{itemize}



We conduct an extensive experimental validation of \ours on 11 few-shot image classification datasets and 8 datasets evaluating robustness. ~We show that \ours outperforms state-of-the-art prompt learning methods on classification accuracy, OOD detection, and domain generalization, therefore improving the observed tradeoff in these 3 criteria. We validate that our two main contributions, \ie learning strong local prompts and diverse representations, are essential for reaching excellent performances.

\section{Related work}\label{sec:related}

\textbf{Prompt learning.} Prompt learning has emerged as an efficient way to adapt VLMs to downstream datasets. These methods, \eg CoOp~\cite{zhou2022learning}, learn \textit{soft prompts} to adapt CLIP textual features to specific labels without the need for a cumbersome step of ``prompt engineering'' as performed in \cite{Radford21}. Following these seminal works, many variants have been proposed. \cite{zhou2022conditional} uses a meta-network to bias the learnable prompt using the global visual representation of the input image. 
To boost prompt learning performances, recent works have focused on learning multiple prompts \cite{Chen22, lu2022prompt, khattak2023maple, khattak2023self}. MaPLe \cite{khattak2023maple} introduces prompts in several layers of both textual and visual encoders. PromptSRC \cite{khattak2023self} builds upon this work by introducing several regularization losses, boosting both accuracy and robustness performances. We note that PromptSRC uses a set of hand-crafted prompts to regularize the learning of the textual prompts, which is not fully aligned with the initial motivation behind prompt learning. Furthermore, both MaPLe and PromptSRC are limited to the use of vision transformer architectures. ProDA \cite{lu2022prompt} models the distribution over the textual representation of classes using a multivariate Gaussian distribution, and indirectly learns the distribution over prompts using a surrogate loss. PromptStyler \cite{ChoICCV23} learns several prompts that represent different ``styles'' to perform source-free domain generalization. These two approaches achieve prompt diversity by enforcing orthogonality among the prompts. In \ours, we induce diversity with a ``prompt dropout'' technique, which randomly drops subsets of prompts during training, thus avoiding the introduction of an additional loss, while limiting prompt over-fitting observed in~\cite{khattak2023self}. We further improve diversity by specializing the local prompts on different image scales to align them with different sets of attributes for each class.

\textbf{Prompt learning using visual local features.} There has been a growing interest in leveraging CLIP's local features in prompt learning methods \cite{Chen22, sun2022dualcoop, Miyai24}.  PLOT \cite{Chen22} learns a set of prompts by using the optimal transport (OT)~\cite{villani2009optimal} distance between them and the set of local features, which is prohibitive to compute. Furthermore, the OT distance enforces the prompts to use information from \emph{all} local visual features during training, including possibly detrimental ones. Also, PLOT adds the global visual features to the local features to achieve strong results on the ImageNet dataset. In \ours,  we use a sparse mechanism to learn localized prompts. This removes the negative influence of background features while being computationally efficient. Finally, \ours learns prompts from the local features without any access to CLIP's original global visual feature. LoCoOp \cite{Miyai24} introduced an entropy loss leveraging ``irrelevant'' local visual features in an outlier exposure fashion \cite{hendrycks2018deep} to improve out-of-distribution detection but at the expense of accuracy. \cite{sun2022dualcoop} introduces a method specifically designed for multi-label classification, which learns prompts using local visual features. While these methods obtained promising results, we show in this work that their performance is intrinsically limited by the lower discriminative power of CLIP's zero-shot local visual features.

\textbf{Prompt learning \& robustness.} 
As VLMs are becoming increasingly prevalent in few-shot classification applications, their robustness capabilities, such are OOD detection or domain generalization (DG), are receiving increasing attention. 
To address OOD detection for VLMs, \cite{ming2022delving} proposed the maximum concept matching (MCM). In \cite{Miyai23}, the authors leveraged  global and local visual information to construct the GL-MCM score to improve OOD detection results. ~In addition to LoCoOp \cite{Miyai24}, other recent works \cite{Nie24} uses negative prompts to perform OOD detection by ``Learning to Say No'' (LSN). Furthermore, prompt learning methods are also evaluated on DG tasks where the prompt is learned on a source dataset (\eg Imagenet) and tested on a domain-shifted target dataset (\eg Imagenet-Sketch). CoOp \cite{zhou2022learning} achieves significantly better DG results than CLIP, and more recent prompt learning methods like \cite{khattak2023self, khattak2023maple} continue to improve over CoOp’s strong performance. By unlocking the potential of local visual features to learn prompts, \ours further improves over the state-of-the-art for top-1 accuracy, OOD detection and domain generalization.

\begin{figure}[t]
    \centering
    \input{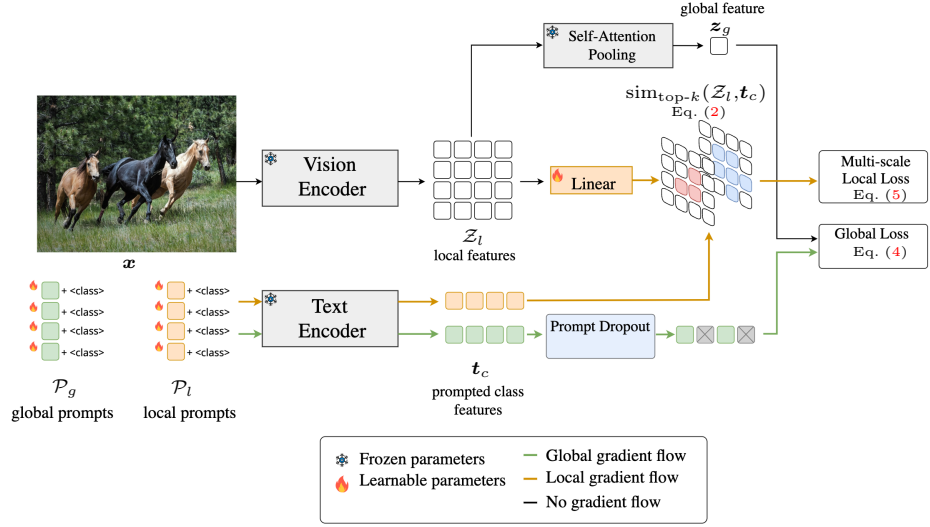}
    \vspace{-1.7em}
   \caption{\textbf{Illustration of \ours.} \ours learns a diverse set of global prompts and local prompts. Pertinent local prompts are learned using only the most relevant regions of the image for each class. We further improve the limited text-vision alignment of CLIP's local features using a simple linear layer. The diversity is encouraged using a new ``prompt dropout'' technique for global prompts, and a multiscale loss for local prompts.} 
   \label{fig:figure_method}
\end{figure}

\section{Combining global and local prompts with \ours}\label{sec:method}

In this section, we describe our proposed method, \ours, which seeks to learn an ensemble of diverse prompts from both global and local CLIP's visual representations. As illustrated in \cref{fig:figure_method}, \ours learns two specialized sets of prompts: the ``global prompts'' receiving a signal from the global visual representation, and the ``local prompts'' trained using local features only.\\

Formally, let us consider a set of $n$ learnable local prompts ${\mathcal{P}_l = (\bm{p}^l_1, \cdots, \bm{p}^l_n)}$ and a set of $m$ learnable global prompts $\mathcal{P}_g = (\bm{p}^g_1, \cdots, \bm{p}^g_m)$. Each of these prompt $\bm{p}$ is composed of $V$ learnable embeddings, \ie $\bm{p} := [p^1, \dots, p^V] \in \mathbb{R}^{V \times d'}$, and are prepended to the class name embeddings $\bm{c}$ to perform classification. 
Let $\mathcal{D} = \{(\bm{x}, ~y)\}$  denote the downstream dataset, where $\bm{x}$ is an image and $y$ its class, and let $\mathcal{T}$ and  $\mathcal{V}$ denote CLIP's text and vision encoder, respectively.  The textual encoder produces a normalized textual representation $\bm{t}_c = \mathcal{T}([\bm{p}, \bm{c}]) \in \mathbb{R}^d$ of the $c^{th}$ class. Given the input image $\bm{x}$, the visual encoder produces a visual representation $\bm{z}$.  $\bm{z}$ can be a global vector for learning global prompts, \ie the global visual feature on which CLIP has been pre-trained. For local prompts, $\bm{z}$ will be a set of localized features outputted by the encoder. From its visual representation $\bm{z}$, the probability for the image $\bm{x}$ to be classified into the class $y_c$ can be expressed as:
\begin{align}\label{eq:promptlearning}
p(y=y_c | \bm{x};~ \bm{p}) &= \frac{\exp(\text{sim}(\bm{z}, ~\bm{t}_c)
 ~/~ \tau)}{\sum_{c'} \exp (\text{sim}(\bm{z}, ~\bm{t}_{c'}) ~/~ \tau)},
\end{align}
where $\text{sim}(\cdot, \cdot)$ is a measure of similarity, and $\tau$ is fixed a temperature scaling parameter. With this general definition of the probability in \cref {eq:promptlearning},  we can train a prompt $\bm{p}$  using the standard cross-entropy loss $\mathcal{L}_{\text{CE}}(p(y=y_c | \bm{x};~ \bm{p}))$.

\vspace{0.5em}
In \cref{sec:method_local}, we introduce a relevant similarity measure $\text{sim}(\bm{z}, ~\bm{t}_c)$ for implementing \cref{eq:promptlearning} on local prompts. We rely on a sparsification strategy that only considers a small subset of class-relevant regions of the image. Furthermore, we use a linear projection to improve the vision-text alignment of local features, thus enhancing the quality of the learned prompts. In \cref{sec:method_diversity} we describe how we learn a diverse set of global and local prompts, whose combination can improve predictions' performance. We introduce ``prompt dropout'' to increase the diversity of global prompts by randomly selecting a subset of prompts for each image. Finally, we introduce a multiscale loss by dedicating each local prompt to select different sub-region sizes of the input image.

\subsection{Learning prompts from local visual representations}\label{sec:method_local}

\begin{wrapfigure}{r}{0.4\linewidth}
   \vspace{-2em}
    \centering
    \input{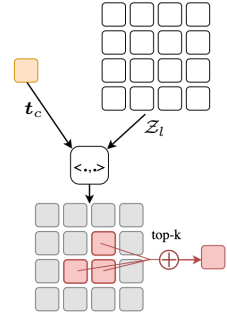}
    \vspace{-0.5em}
    \caption{\textbf{\ours sparse local similarity} $\text{sim}(\mathcal{Z}_l,~ \bm{t}_c)$ between class prompt $\bm{t}_c$ and visual features $\mathcal{Z}_l$ is the average of the top-$k$ highest similarities (here, k=3).}
    \vspace{-2em}
\label{fig:sparsity}
\end{wrapfigure}

In this section, we temporarily consider a single local prompt $\bm{p}^l_j \in \mathcal{P}_l$ without loss of generality.  In this case, the visual representation $\bm{z}$ that we consider is the set of visual local features, \ie  $ \bm{z} = \mathcal{Z}_l\in \mathbb{R}^{L \times d}$, obtained following~\cite{dong2023maskclip} (see details in supplementary~\cref{supp:clip_local}). Here, we can not directly compute the probability of \cref{eq:promptlearning} as we need to define the similarity between the set of vectors $\mathcal{Z}_l = (\bm{z}^l_1, \cdots, \bm{z}^l_L)$ and the textual representation of the $c^{th}$ class, $\bm{t}_c = \mathcal{T}([\bm{p}^l_j, \bm{c}])$.

\vspace{0.7em}
\noindent\textbf{Sparse local similarity.} A naive way to obtain a single similarity for all regions is to average the similarities of each spatial location with the textual representation of the class. However, a substantial portion of the local features are irrelevant to the class, \eg features from background areas, which may introduce noise and perturb the learning process. To solve this problem, we adopt a sparse approach, where only local features semantically related to the class are kept to perform classification. As illustrated in \cref{fig:sparsity}, we select the top-$k$ local features with the highest similarities with the prompted class textual representation, and average their similarities to measure $\text{sim}(\mathcal{Z}_l,~ \bm{t}_c)$.

Formally, we define the similarity between a prompt $\bm{t}_c$ and the set of visual features $\mathcal{Z}_l$ as the average similarity for the $k$ most similar regions:
\begin{align}\label{eq:sparsity}
     \text{sim}_{\text{top-$k$}}(\mathcal{Z}_l,~ \bm{t}_c)& \coloneqq \frac{1}{k} \sum_{i=1}^L ~\mathbb{1}_{\text{top-$k$}}(i) \cdot \left<\bm{z}^l_i, ~\bm{t}_c\right> \\
    \text{where} \quad \quad\mathbb{1}_{\text{top-$k$}}(i) &= \left\{\begin{array}{ll}
        1 & \quad\text{if} \quad \text{rank}_i(\left<\bm{z}^l_i, ~\bm{t}_c\right>) \leq k, \\
        0 & \quad \text{otherwise.}
    \end{array}
    \right. \nonumber
\end{align}

\noindent which we plug into \cref{eq:promptlearning} to compute the probability for class $c$.
We show in~\cref{sec:ablation_studies} that relying on sparsity is mandatory for local prompt learning, boosting performances by almost 20pt in top-1 accuracy.

\vspace{0.5em}
\noindent\textbf{Improving local text-vision alignment.}
While previous works \cite{ZhouLD22, sun2022dualcoop, Miyai24} have exploited the text-vision alignment of CLIP's local features, we empirically verified in \cref{sec:ablation_studies} that using these features leads to poor zero-shots classification results on ImageNet. This is expected, as CLIP is pre-trained to align the global visual features with its textual representation. Local features are thus suboptimal to learn effective prompts for image classification. Motivated by this observation, we propose to improve the discriminative power of CLIP's local visual features by realigning them with the textual representations of the class labels of the downstream dataset. To do so, we propose to use a simple linear projection $h_{\bm{\theta}}$. To ease the learning process, we initialize the linear layer $h_\theta$ to identity, so that the initial features are close to CLIP's representations. Henceforth, we use the set of linearly transformed local visual features $h_{\bm{\theta}}(\mathcal{Z}_l)$ to compute the probability of \cref{eq:promptlearning}, which becomes:
\vspace{-0.5em}
\begin{align}\label{eq:proba_aligned}
p(y=y_c | \bm{x};~ \bm{p}^l_j, k, \bm{\theta}) &= \frac{\exp( \text{sim}_{\text{top-$k$}}(h_{\bm{\theta}}(\mathcal{Z}_l),~ \bm{t}_c)
 ~/~ \tau)}{\sum_{c'} \exp (\text{sim}_{\text{top-$k$}}(h_{\bm{\theta}}(\mathcal{Z}_l),~ \bm{t}_{c'}) ~/~ \tau)}.
\end{align}

\noindent Thus, a local prompt can be optimized by maximizing this probability with the cross-entropy loss. These design choices in \ours allow us to train a powerful classifier for local features: the sparsity helps to focus on the most relevant regions of an image and to remove potential background noise, while the linear projection enhances the text-vision alignment and boosts the fine-grained discriminating power of the local features. We study these design choices in~\cref{sec:ablation_studies}.

\subsection{Learning multiple diverse prompts} \label{sec:method_diversity}

In this section, we describe how we induce diversity among the learned prompts. Besides exploiting different sources of information -- the global and visual ones --,  we introduce two mechanisms to increase diversity: ``prompt dropout'' and multiscale training. 

To train our set of global prompts, we can simply train each one of them independently using the cross-entropy loss as in CoOp \cite{zhou2022learning}. However, this strategy will necessarily produce identical global prompts, losing the advantage of using an ensemble of prompts. A possible strategy to avoid this behavior is to use an explicit diversity loss inducing a semantic orthogonality between the different global prompts. Instead of adding a another loss term, which can disturb the training process, we chose to take a different approach.

\begin{figure}[t]
\begin{tabular}{cccc}
\hspace{0.05\textwidth} & \includegraphics[width=0.4\textwidth]{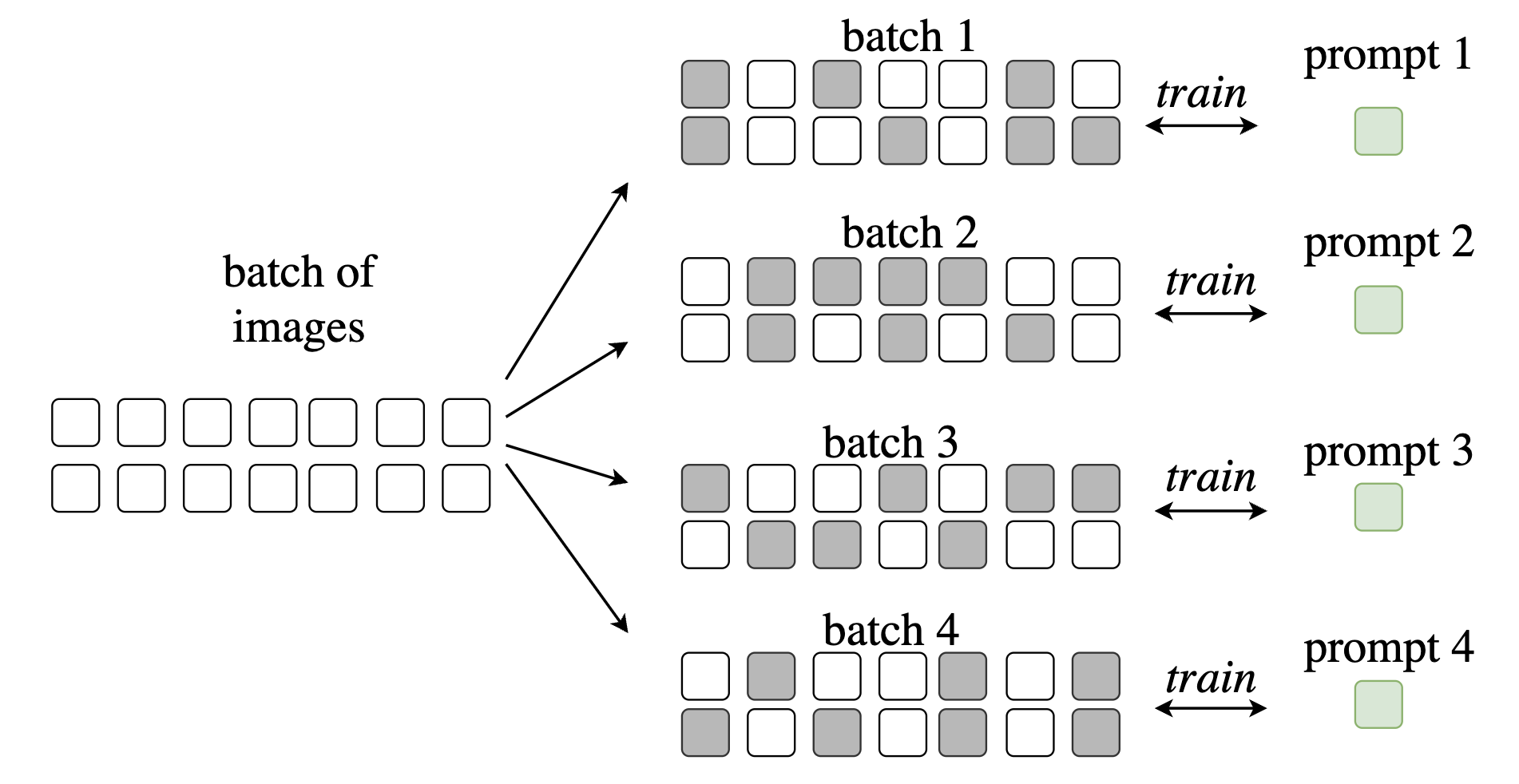} & \hspace{0.075\textwidth} & \includegraphics[width=0.4\textwidth]{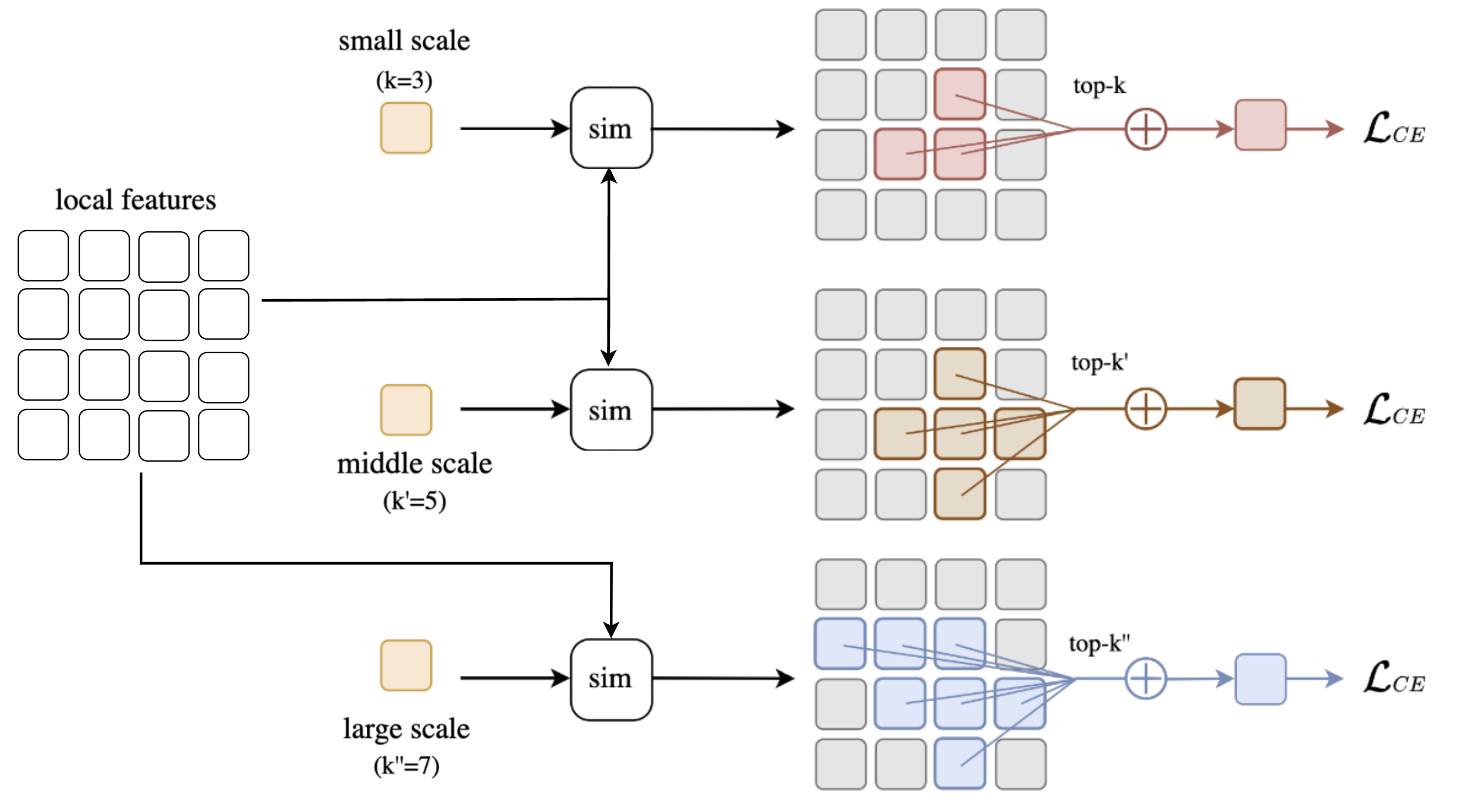}\\
& (a) Prompt dropout & & (b) Multiscale loss 
\end{tabular}
\vspace{-0.5em}
\caption{(a) Prompt dropout induces diversity by randomly selecting different subsets of prompts for each image of the batch. In (a), each image will be used by half the prompts. (b) To learn diverse local prompts, we specialize each one of them using a different number of regions, and therefore a different level of sparsity.}
\label{fig:diversity}
\vspace{-0.5em}
\end{figure}

\vspace{0.5em}
\noindent\textbf{Prompt dropout.}  Motivated by the success of the ``dropout'' \cite{srivastava2014dropout, gal2016dropout} technique classically used in deep learning, we introduce ``prompt dropout'' into the prompt learning framework. In ``prompt dropout'', we randomly mask a subset of prompts for each image of the batch. Alternatively, from the perspective of each prompt, we select a different subset of the batch of images, thus inducing diversity in the learning process of the prompts through input randomization (see \cref{fig:diversity}(a)).
Formally, the loss used to train our global prompts $\mathcal{P}_g$ with prompt dropout can be expressed as:
\begin{align}\label{eq:global_loss}
    \mathcal{L}_{\text{global}}(\mathcal{P}_g) &=~\mathbb{E}_{\bm{x},y}\sum_{i = 1}^m \delta_i(\bm{x})\mathcal{L}_{\text{CE}}(p(y|\bm{x};~ \bm{p}^g_i))
\end{align}
where $\delta_i(\bm{x})\sim\text{Bernoulli}(1-r)$ with $r$ the dropout rate.

\vspace{0.5em}
\noindent\textbf{Multiscale training.} To specifically improve the diversity of the local prompts, we specialize each local prompt to select a different number of class-specific visual patches (scales). In this way, prompts dedicated to small scales will get more signals from classes corresponding to small visual concepts, \eg ``daisy flower'' or ``tailed frog'', while prompts learned with larger scales will receive more signals from images with wider concepts, \eg ``castle'' or ``valley''. More formally, let $(k_1,~  k_1 + \Delta_k, \cdots,~  k_1 + (n-1)\cdot \Delta_k)$ denote a set of increasing scales with $k_1$ the first scale and $\Delta_k$ the expansion factor. Each local prompt $\bm{p}^l_j$ will be learned with its associated scale $k_j= k_1 + (j-1)\cdot \Delta_k$.\\

The training of our $n$ local prompts is performed by optimizing the probability defined in~\cref{eq:proba_aligned} for each prompt with a different scale, \ie value of $k$:
\begin{align}\label{eq:multiscale_loss}
    \mathcal{L}_{\text{x-scale}}(\mathcal{P}_l, \bm{\theta}) ~=~ ~\mathbb{E}_{\bm{x},y} \sum_{j = 1}^n \mathcal{L}_{\text{CE}}(p(y|\bm{x};~ \bm{p}^l_j,~ \bm{\theta}, ~k_j)).
\end{align}

\noindent The overall loss to train our set of prompts $\mathcal{P} = \mathcal{P}_l \cup \mathcal{P}_g$ is the sum of the local multiscale and global losses:
\begin{align}\label{eq:total_loss}
    \mathcal{L}_{\text{total}}(\mathcal{P}, \bm{\theta}) = \mathcal{L}_{\text{global}}(\mathcal{P}_g) + \mathcal{L}_{\text{x-scale}}(\mathcal{P}_l, \bm{\theta})
\end{align}

\paragraph{Inference.} We perform averaging of the similarities obtained with each prompt to obtain a final similarity, $\text{sim}(\bm{z}, ~\bm{t}_c)$, for each class 
 which is used as logit for classification. For OOD detection we use the GL-MCM score \cite{Miyai23}. The inference procedures are detailed in \cref{supp:inference}.
\newpage
\section{Experimental results}\label{sec:expes}

In this section, we present the experimental validation of \ours. We first show, in~\cref{sec:top1_performances}, that \ours outperforms previous methods on top-1 accuracy on a collection of 11 datasets used in~\cite{zhou2022learning} with ViT-B/16~\cite{dosovitskiy2020image}. We also show that \ours performs well for different few shot settings on ImageNet and with a ResNet-50~\cite{he2015deep}. In~\cref{sec:robustness_results}, we compare robustness performances of \ours and other prompts learning methods in domain generalization and OOD detection, and show that \ours has better trade-off with top-1 accuracy contrary to previous methods. In~\cref{sec:ablation_studies}, we conduct ablation studies of the different components of \ours.

\begin{table}[b]
    \setlength\tabcolsep{0.5pt}
    \caption{Top-1 accuracy with ViT-B/16 backbone. Comparison of \ours to other prompt learning methods on several standard benchmarks. $^\dagger$results based on our own re-implementation.}
    \vspace{-1em}
    \label{tab:top1_accuracy_vit}
    \centering
    \begin{tabularx}{\linewidth}{l YYYYYYYYYYYYY}
        \toprule
         Dataset & \rotatebox[origin=c]{70}{ImageNet~\cite{deng2009imagenet}} & \rotatebox[origin=c]{70}{Caltech101~\cite{FeiFei2004LearningGV}} & \rotatebox[origin=c]{70}{OxfordPets~\cite{parkhi12a}} & \rotatebox[origin=c]{70}{Cars~\cite{KrauseStarkDengFei-Fei_3DRR2013}} & \rotatebox[origin=c]{70}{Flowers102~\cite{Nilsback08}} & \rotatebox[origin=c]{70}{Food101~\cite{bossard14}} & \rotatebox[origin=c]{70}{Aircraft~\cite{maji13fine-grained}} & \rotatebox[origin=c]{70}{SUN397~\cite{Xiao:2010}} & \rotatebox[origin=c]{70}{DTD~\cite{cimpoi14describing}} & \rotatebox[origin=c]{70}{EuroSAT~\cite{helber2017eurosat}} & \rotatebox[origin=c]{70}{UCF101~\cite{soomro2012ucf101}} & \rotatebox[origin=c]{70}{Average} \\
         \midrule
         CLIP~\cite{Radford21} & 66.7 & 92.2 & 88.4 & 65.5 & 70.7 & 84.8 & 24.8 & 62.3 & 44.1 & 48.3 & 64.7 & 75.7\\
         Linear Probe & 67.3 & 95.4 & 85.3 & 80.4 & 97.4 & 82.9 & 45.4 & 73.3 & 70.0 & 87.2 & 82.1 & 78.8\\
         CoOp~\cite{zhou2022learning} & 71.7 & 95.6 & 91.9 & 83.1 & 97.1 & 84.2 & 43.4 & 74.7 & 69.9 & 84.9 & 82.2 & 79.9\\
         Co-CoOp~\cite{zhou2022conditional} & 71.0 & 95.2 & 93.3 & 71.6 & 87.8 & \textbf{87.2} & 31.2 & 72.2 & 63.0 & 73.3 & 78.1 &  74.9\\
         MaPLe \cite{khattak2023maple} & 72.3 & \underline{96.0} & 92.8 & 83.6 & 97.0 & 85.3 & 48.4 & 75.5 & 71.3 & \textbf{92.3} & 85.0 & 81.8\\
         PLOT \cite{Chen22} & 72.6 & \underline{96.0} & \underline{93.6} & 84.6 & \underline{97.6} & \textbf{87.1} & 46.7 & 76.0 & 71.4 & 92.0 & 85.3 & 82.1\\
         PromptSRC$^\triangleright$ \cite{khattak2023self} & \underline{73.2} &  \underline{96.1} & \underline{93.7} & \underline{85.8} & \underline{97.6} & \underline{86.5} & \underline{50.8} & \textbf{77.2} & \underline{72.7} & \textbf{92.4} & \underline{86.5} & \underline{82.9}\\
         LoCoOp$^\dagger$ \cite{Miyai24} & 71.5 & 94.9 & 92.4 & 79.8 & 96.3 & 84.7 & 40.7 & 74.2 & 69.5 & 86.1 & 81.6 & 79.2\\
         ProDA$^\dagger$ \cite{lu2022prompt} & 71.9 & 95.5 & 93.5 & 79.8 & 96.8 & 86.8& 40.2 & 75.7 & 70.9 & 85.1 & 83.3 & 80.0\\[2pt]
         \rowcolor{mycornflowerblue} \textbf{\ours} & \textbf{75.1} & \textbf{96.7} & \textbf{94.1} & \textbf{89.2} & \textbf{98.8} & \underline{86.5} & \textbf{58.3} & \textbf{77.2} & \textbf{75.5} & \underline{90.1} & \textbf{86.9} & \textbf{84.4}\\
         \bottomrule
    \end{tabularx}
\end{table}

\paragraph*{Implementation details.} We experiment with both ResNet-50 and ViT-B/16 CLIP models. When not specified, we use ViT-B/16.
We train for 50 epochs on ImageNet and 200 epochs for other datasets with SGD, a learning rate of 0.002 decayed using cosine annealing and a weight decay of 0.01, following the setting of \cite{zhou2022learning}. Unless specified otherwise, we train the models using 16 shots. Our base parameters for \ours are as follows: $m=4$ global prompts with a dropout of $r=75\%$ (in practice we keep a single prompt for each image), $n=4$ local prompts with scales $k_1 = 10$ and $\Delta_k = 10$ for ViT-B/16 and $k_1 = 5$ and $\Delta_k = 5$ for ResNet-50 as there are fewer local patches. We keep $\tau$ fixed from CLIP.

\paragraph*{Baselines.} We compare \ours to recent prompt learning methods. Including, single prompt learning CoOp and Co-CoOp. Multi-prompt learning MaPLe, ProDA, PLOT, PromptSRC. We denote by PromptSRC$^\triangleright$ the version designed for accuracy and PromptSRC$^\diamond$ the version designed for domain generalization. We also include OOD detection specific methods such as LoCoOp and LSN.

\subsection{Main in-distribution results.}\label{sec:top1_performances}

On~\cref{tab:top1_accuracy_vit}, we compare \ours with a ViT-B/16 backbone on a suite of 11 datasets, a standard benchmark for prompt learning methods. On average, \ours outperforms previous methods by a large margin with +1.5pt compared to PromptSRC$^\triangleright$ the next best performing method. Furthermore, \ours performs well on most datasets, achieving state-of-the-art among prompt learning methods. For instance, on the large-scale ImageNet dataset, it outperforms PLOT by +2.5pt and PromptSRC$^\triangleright$ by +1.9pt. On some datasets, \eg FGVC Aircraft, \ours outperforms the next best method by a large margin, with +7.5pt compared to PromptSRC$^\triangleright$. 

\begin{figure}[t]
    \centering
    \begin{subfigure}[t]{0.49\linewidth}
        \vspace{0pt}
        \centering
        \includegraphics[width=\linewidth]{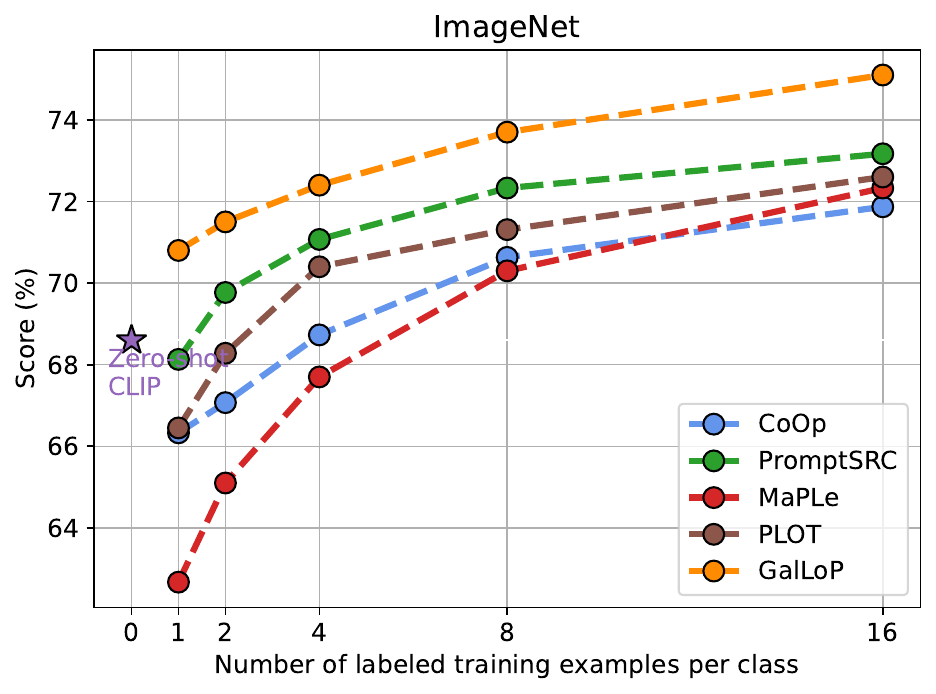}
        \caption{Few shot results.}
        \label{fig:main_few_shot_imagenet}
    \end{subfigure}
    \hfill
    \begin{subfigure}[t]{0.49\linewidth}
        \vspace{0pt}
        \centering
        \includegraphics[width=\linewidth]{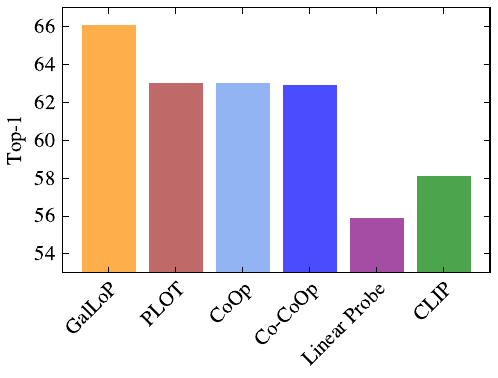}
        \caption{Results with ResNet-50.}
        \label{fig:backbone_bar_plot}
    \end{subfigure}
    \caption{Results on ImageNet with different few shot settings~\cref{fig:main_few_shot_imagenet}, and ResNet-50~\cref{fig:backbone_bar_plot}.}
    \label{fig:additional_imagenet_results}
\end{figure}

We then compare \ours on~\cref{fig:main_few_shot_imagenet} to prompt learning methods in different few-shot settings on ImageNet. \ours performs well in all configurations, outperforming for each setting the very competitive method, PromptSRC$^\triangleright$. Finally, in~\cref{fig:backbone_bar_plot} we show that \ours works well with a ResNet-50, outperforming PLOT and CoOp by +3.1pt. Note that compared to other methods, \eg MaPLe and PromptSRC, \ours is amenable to both convolutional and transformer vision backbones. \textcolor{black}{Detailed results for ResNet-50 can be found in the supplementary material \ref{sec:backbones}}

\subsection{Robustness results.}\label{sec:robustness_results}

\begin{figure}[h]
    \centering
    \begin{subfigure}[t]{0.45\linewidth}
        \vspace{0pt}
        \centering
        \includegraphics[width=\linewidth]{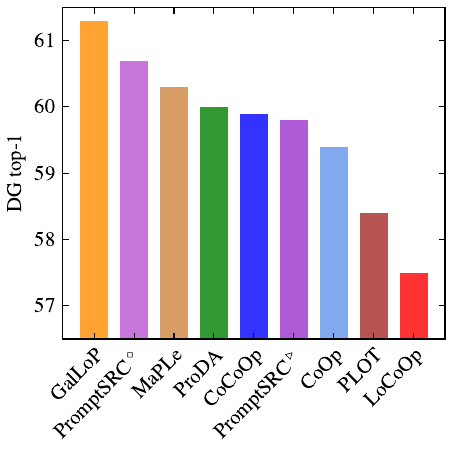}
        \caption{Domain generalization.}
        \label{fig:DG_Average}
    \end{subfigure}
    \hspace{1em}
    \begin{subfigure}[t]{0.45\linewidth}
        \vspace{0pt}
        \centering
        \includegraphics[width=\linewidth]{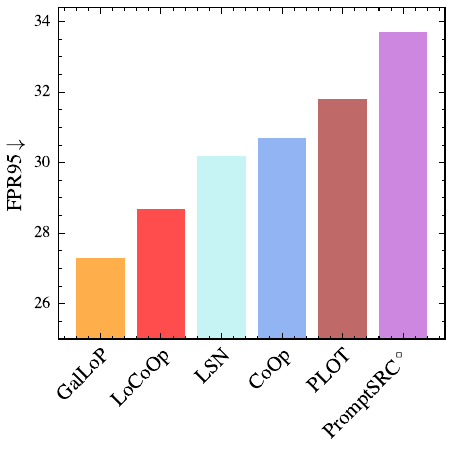}
        \caption{OOD detection.}
        \label{fig:ood_FPR_Average}
    \end{subfigure}
    \caption{\textbf{\ours robustness performances}. \ours achieves strong performances on domain generalization \cref{fig:DG_Average} and on OOD detection~\cref{fig:ood_FPR_Average} ImageNet benchmarks, while outperforming prompt learning methods on top-1 accuracy (detailed results in \cref{tab:domain_generalization_vit} and \cref{tab:ood_detection_vit}).}
    \label{fig:robustness}
\end{figure}

In this section, we compare the robustness performances of \ours \vs other prompt learning methods, see~\cref{fig:robustness}, on domain generalization and OOD detection. For both benchmarks, models are trained on ImageNet (16 shots).

\vspace{0.5em}
\noindent\textbf{Domain generalization results.}\label{sec:domain_generalizarion}
We compare on~\cref{fig:DG_Average} the domain generalization performances of \ours \vs other prompt learning methods. After being trained on ImageNet (16 shots), the models are evaluated on top-1 accuracy for different domains with the same classes as ImageNet, \ie ImageNet-V2~\cite{recht2019imagenet}, ImageNet-Sketch~\cite{wang2019learning}, ImageNet-A~\cite{hendrycks2021nae} and ImageNet-R~\cite{hendrycks2021many}. \ours outperforms the domain-generalization specific method PromptSRC$^\diamond$ by +0.5pt on average, while outperforming it by +4.9pt on ImageNet. This illustrates the trade-off made by PromptSRC between top-1 accuracy and domain generalization. Indeed, \ours outperforms PromptSRC$^\triangleright$, designed for ImageNet accuracy, by +1.9pt on ImageNet and +1.5pt on average in domain generalization. \ours achieves the best trade-off between top-1 performances and domain generalization. \textcolor{black}{The detailed results can be found in supplementary material \ref{sec:sup_dg}}



\vspace{0.5em}
\noindent\textbf{Results on OOD detection.}\label{sec:exp_ood_detection} 
In OOD detection the models must recognize between in-distribution examples (ImageNet test set) and different OOD datasets, namely iNaturalist~\cite{van2018inaturalist}, SUN~\cite{Xiao:2010}, Places~\cite{zhou2017places} and Textures~\cite{cimpoi14describing}, a standard benchmark in the OOD detection literature. We plot on~\cref{fig:ood_FPR_Average} the average results on the ImageNet OOD benchmark of \ours and other prompt learning methods measured in FPR95 (lower is better, $\downarrow$). \ours outperforms traditional prompt learning methods, \eg CoOp -3pt FPR95, as well as dedicated OOD detection methods, \eg -1.4pt FPR95 \vs LoCoOp or -2.9pt FPR95 \vs LSN. Meanwhile, \ours also outperforms both LSN and LoCoOp by a large margin in top-1 accuracy, \ie +3.2pt and +3.6pt respectively. \textcolor{black}{The detailed results can be found in the supplementary material \ref{sec:sup_ood}}

\subsection{Ablation studies.}\label{sec:ablation_studies}

\begin{wrapfigure}{r}{0.45\linewidth}
\vspace{-1.2em}
    \centering
    \includegraphics[width=\linewidth]{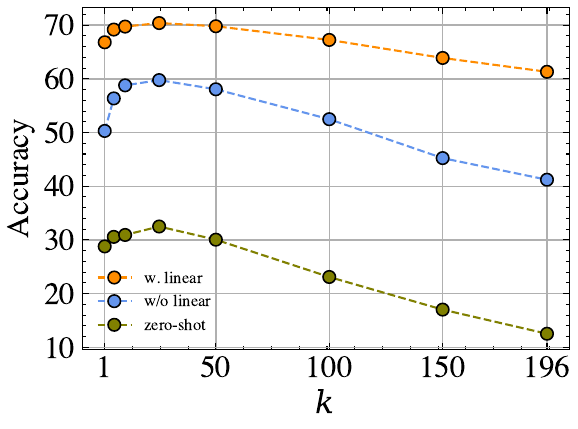}
    \caption{Impact of our sparsity choice for three regimes, zero-shot CLIP, learning a local prompt (``w/o linear'') and aligning our linear projection with a local prompt (``w. linear'') (ImageNet, 16 shots).}
    \label{fig:abaltion_sparsity}
    \vspace{-1.5em}
\end{wrapfigure}

In this section, we investigate -- on ImageNet 16 shots -- the design choices for \ours. We first show how \ours leverages the complementarity of strong global and local prompts to boost performances~\cref{tab:ablation_local_global}. We then demonstrate the benefit of sparsity and local alignment in~\cref{fig:abaltion_sparsity}. Finally, we show the impact of our choice when learning multiple prompts for both global and local features~\cref{fig:diversity_ablation}.

\begin{wraptable}{r}{0.5\linewidth}
    \vspace{-1em}
    \setlength\tabcolsep{0.7pt}
    \caption{Ablation studies for the different components of our \ours (ImageNet, 16 shots).}
    \centering
    \begin{tabularx}{\linewidth}{l YYY r}
        \toprule
          & \small{Top-1} & \small{DG} & \small{FPR95}\scriptsize{$\downarrow$}~ & ~~\small{AUC} \\
         \midrule
         CLIP\textsubscript{Global} & 66.6 & 57.2 & 42.8 & 90.8 \\
         CLIP\textsubscript{Local} & 12.5 & 9.49 & 73.3 & 73.7\\
         CLIP\textsubscript{GL} & 61.1 & 49.3 & 35.5 & 90.8 \\
         \midrule
         CoOp\textsubscript{Global} & 71.4 & 59.2 & 39.1 & 91.1 \\
         CoOp\textsubscript{Local} & 41.2 & 30.1  & 65.2 & 78.3 \\
         CoOp\textsubscript{GL} & 69.5 & 55.6 & 33.7 & 90.5 \\
         \midrule
         GalLoP\textsubscript{Global} & 72.0 & 60.4 & 37.0 & 91.7\\
        GalLoP\textsubscript{Local} & 70.9 & 54.1  & 36.0 & 90.1\\
         \rowcolor{mycornflowerblue} \ours & \textbf{75.1} & \textbf{61.3} & \textbf{27.3} & \textbf{93.2} \\
         \bottomrule
    \end{tabularx}
    \label{tab:ablation_local_global}
    \vspace{-1.2em}
\end{wraptable}

\noindent\textbf{Combining global and local features.} On~\cref{tab:ablation_local_global}, we show that leveraging global and local features requires some important design choices. Indeed, we experiment with a baseline using CoOp on local features (``CoOp\textsubscript{Local}''), learning a single prompt, without sparsity and no alignment. This baseline already outperforms using zero-shot local features, +28.7pt top-1. However, its combination with a standard CoOp\textsubscript{Global}, \ie ``CoOp\textsubscript{GL}'', is detrimental to final top-1 performances, with -1.9pt top-1 or -3.6pt DG compared to CoOp\textsubscript{Global}. On the other hand, \ours enjoys a boost in performances on all metrics when combining the learned global (GalLoP\textsubscript{Global}) and local (GalLoP\textsubscript{Local}) prompts. We can see that the top-1 performances of \ours increase by +3.1pt compared to (GalLoP\textsubscript{Global}). Similarly, on OOD detection, \ours has a decrease of -8.9pt FPR95 compared to GalLoP\textsubscript{Local}. \cref{tab:ablation_local_global} illustrates how the resulting performances of \ours, in both accuracy and robustness, come from the complementarity of both the local and global features.

\vspace{0pt}
\noindent\textbf{The need for sparsity.} In~\cref{fig:abaltion_sparsity} we show how the sparsity when using local features allows achieving higher performances than attending to each local feature, for three regimes: zero-shot CLIP (``zero-shot''), while learning a local prompt (``w/o linear''), and when aligning a local prompt and our linear projection (``w. linear''). On the three regimes, the difference between looking at all local features and the best reported sparsity level is, respectively, +18.4pt, +17.6pt, and +8.5pt. Furthermore, we can see that when aligning a local prompt and the linear layer, our sparsity ratio works for a wide range of $k$, with performances above 69pt between $k=5$ and $k=50$. This shows the robustness to the choice of $k$. Finally, learning a local prompt allows to significantly boost the performances for the local features, \eg +27.9pt for $k=10$, and aligning with a linear projection further boosts performances, with +10pt for $k=10$ compared to learning the prompt only. \cref{fig:abaltion_sparsity} shows the interest of both enforcing the sparsity when looking at local features and further aligning the local features with a local prompt.

\vspace{0.5em}
\noindent\textbf{Global prompt learning with prompt dropout.} We display on~\cref{fig:ablation_prompt_dropout} how prompt dropout allows learning efficiently multiple prompts for the global features. We display the top-1 accuracy when using more and more prompts, with (``w.'') or without (``w/o'') prompt dropout. We can observe that adding more prompts does not result in better performances without prompt dropout. For example, performances with 6 prompts decrease compared to using a single prompt . This is due to limited diversity among the learned prompts. In comparison, adding more prompts is always beneficial when using prompt dropout.

\newcommand\AblplotSize{0.4}
\begin{figure}[t]
    \centering
    \begin{subfigure}[t]{\AblplotSize\linewidth}
        \vspace{0pt}
        \centering
        \includegraphics[width=\linewidth]{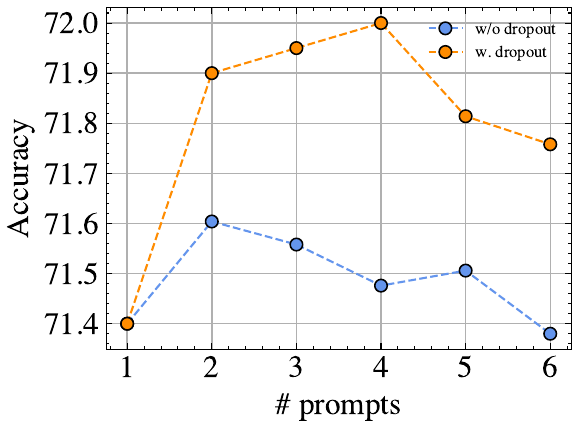}
        \caption{Impact of prompt dropout when learning multiple global prompts.}
        \label{fig:ablation_prompt_dropout}
    \end{subfigure}
    \hspace{2em}
    \begin{subfigure}[t]{\AblplotSize\linewidth}
        \vspace{0pt}
        \centering
        \includegraphics[width=\linewidth]{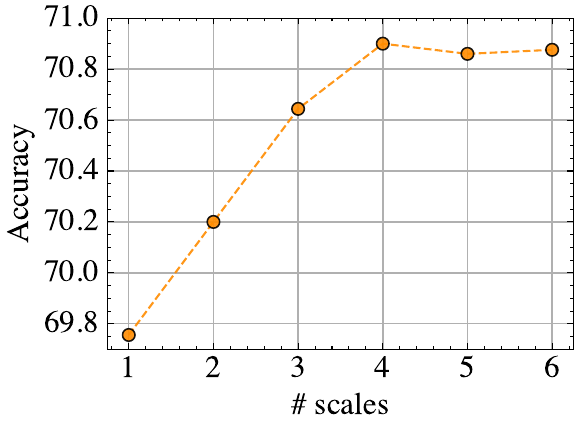}
        \caption{Impact of multiple scales when learning local prompts, with $k_1 = 10$, $\Delta_k = 10$.}
        \label{fig:abaltion_multiscale}
    \end{subfigure}
    \caption{Impact of our design choices on learning 
    global~\cref{fig:ablation_prompt_dropout}
    and local prompts~\cref{fig:abaltion_multiscale}
    .}
    \label{fig:diversity_ablation}
\end{figure}

\noindent\textbf{Local prompt learning at multiple scales.} On~\cref{fig:abaltion_multiscale}, we show the interest of our multiscale approach. We experiment with various number of scales, \ie from 1 to 6 scales with $k_1 = 10$ and $\Delta_k = 10$ and report the top-1 accuracy. We can observe a steady increase from 1 scale to 4 scales (+1pt). Performances stabilize afterward for 5 and 6 scales. \cref{fig:abaltion_multiscale} shows that learning at different scales is beneficial, but also that \ours is not too sensitive to the choice of number of prompts. Furthermore, learning at different scale also reduce the need to select an optimal $k$, although we show in~\cref{fig:abaltion_sparsity} that performances are stable with respect to $k$.

\subsection{Qualitative study.}\label{sec:qualitative_study}

\begin{figure}[h]
\centering
\begin{tabular}{c}
 \includegraphics[width=\linewidth]{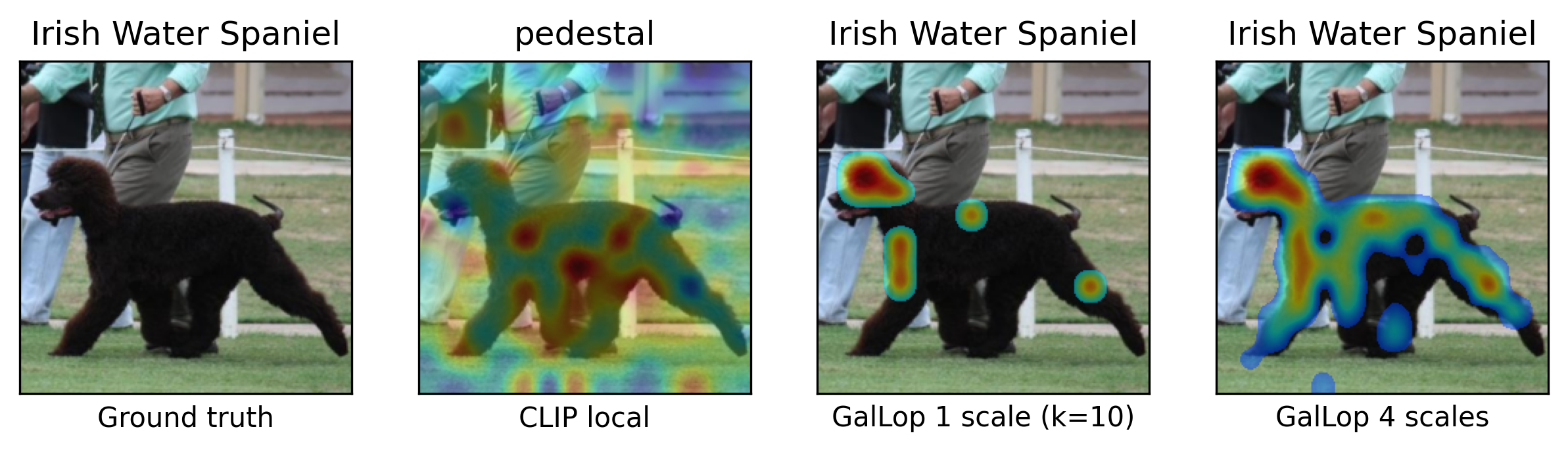} \\
 \includegraphics[width=\linewidth]{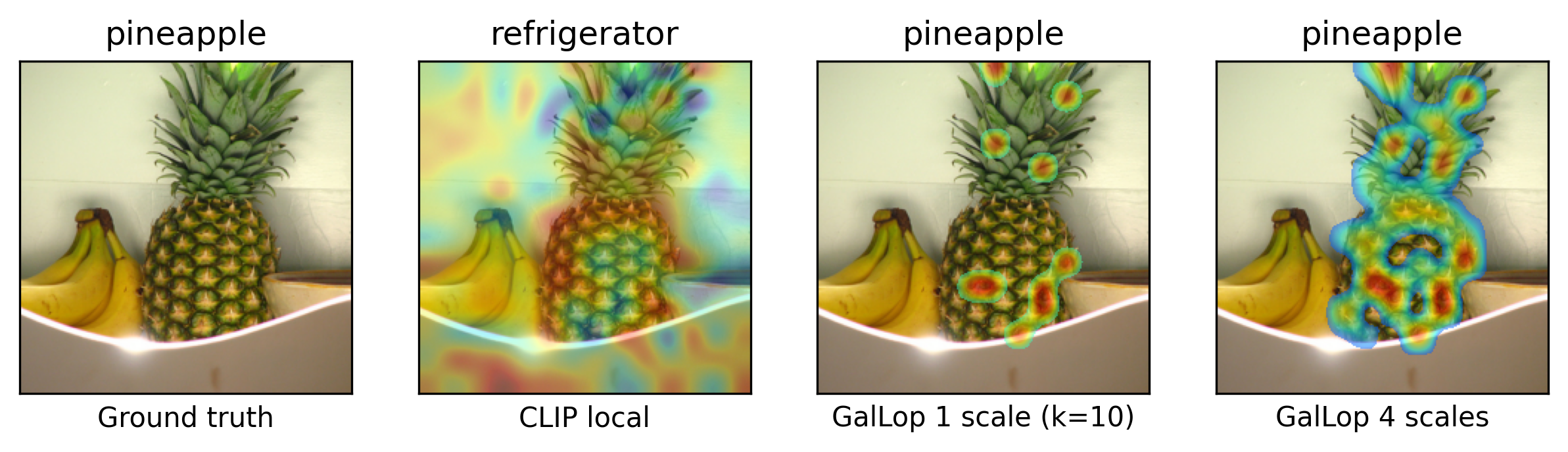} \\
\end{tabular}
\vspace{-1.5em}
\caption{\textbf{Qualitative comparison of CLIP and \ours.} From left to right, the original image with its ground truth, CLIP local wrong prediction, one scale ($k$=10) of \ours with correct prediction and \ours multiscale, resulting in correct prediction and segmentation.}
\label{fig:qual_results_compa_clip}
\vspace{-1.5em}
\end{figure}

We conduct in this section a qualitative study of \ours, by comparing it to CLIP on~\cref{fig:qual_results_compa_clip}, and visualizing its different scales on~\cref{fig:qual_results_multiscale}. \textcolor{black}{We show other qualitative results in supplementary material \ref{sec:sup_quali}}

\vspace{0.5em}
\noindent\textbf{Comparison to CLIP.} On~\cref{fig:qual_results_compa_clip}, we compare \ours and CLIP local features. We can observe that CLIP's local features are not discriminative and do not allow to classify images correctly, which was observed in~\cref{sec:ablation_studies}. On the other hand, \ours classifies correctly the images, even with a single scale. We can also observe \ours accurately segments the object of interest when using all its scales.

\vspace{0.5em}
\noindent\textbf{Visualize multiple scales.} Finally, we show the different regions each of the local prompts attend to. We can see that scale \# 1 focuses on the most discriminative features, \ie the head and tail of the ``Ring tailed lemur''. Each scale progressively attends to different parts of the body, leading to an accurate prediction.

\begin{figure}[h]
\centering
 \includegraphics[width=\linewidth]{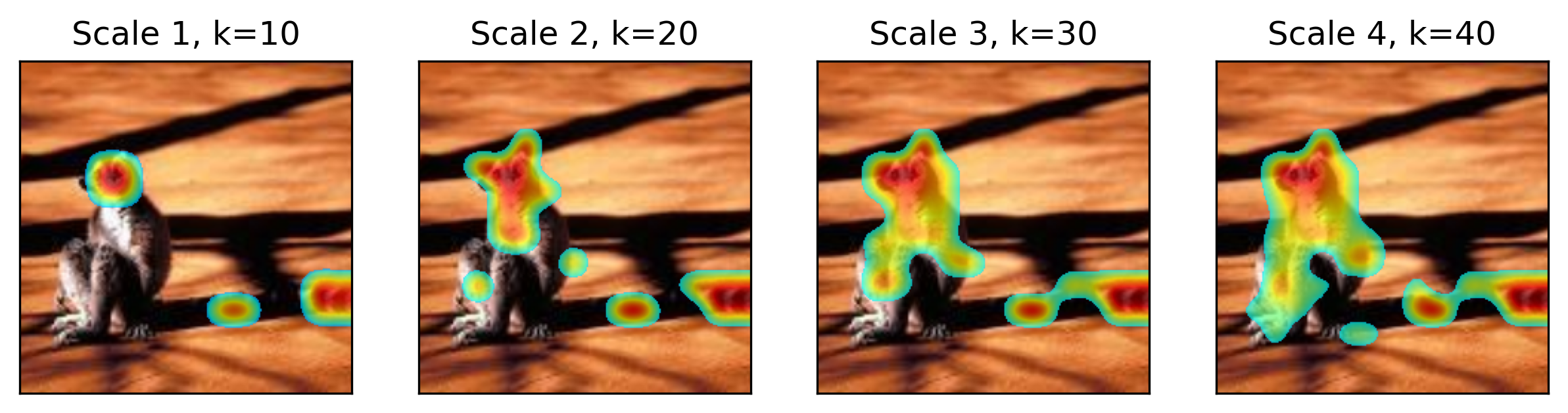} \\
\vspace{-1em}
\caption{\textbf{\ours multiscale visualization}. Regions observed by the different prompts of \ours for a ``Ring tailed lemur''.}
\label{fig:qual_results_multiscale}
\end{figure}

\section{Conclusion}\label{sec:conclusion}

This paper introduces \ours, a new prompt learning method that leverage both global and local visual representations. The key features of \ours are the strong discriminability of its local representations and its capacity to produce diverse predictions from both local and global prompts. ~Extensive experiments show that \ours outperforms previous prompt learning methods on top-1 accuracy on average for 11 datasets; that it works in different few shot settings; and for both convolutional and transformer vision-backbones. We show in ablation studies the interest of the design choices that make \ours work, \ie complementarity between local and global prompts; sparsity and enhanced alignment; encouraging diversity. Finally, we conduct a qualitative study to show what local prompts focus on when classifying an image. Future works include learning the local feature alignment on a large vision-language dataset.

\section*{Acknowledgements}
This work was done under grants from the DIAMELEX ANR program (\textcolor{black}{ANR-20-CE45-0026}) and the AHEAD ANR program (\textcolor{black}{ANR-20-THIA-0002}). It was granted access to the HPC resources of IDRIS under the allocation AD011012645R1 and AD011013370R1 made by GENCI.


%
%
\bibliographystyle{splncs04}
\bibliography{main}

\begin{thebibliography}{10}
\providecommand{\url}[1]{\texttt{#1}}
\providecommand{\urlprefix}{URL }
\providecommand{\doi}[1]{https://doi.org/#1}

\bibitem{agnolucci2023eco}
Agnolucci, L., Baldrati, A., Todino, F., Becattini, F., Bertini, M., Del~Bimbo, A.: Eco: Ensembling context optimization for vision-language models. In: ICCV (2023)

\bibitem{bossard14}
Bossard, L., Guillaumin, M., Van~Gool, L.: Food-101 -- mining discriminative components with random forests. In: ECCV (2014)

\bibitem{Chen22}
Chen, G., Yao, W., Song, X., Li, X., Rao, Y., Zhang, K.: Plot: Prompt learning with optimal transport for vision-language models. In: The Eleventh International Conference on Learning Representations (2023)

\bibitem{ChoICCV23}
Cho, J., Nam, G., Kim, S., Yang, H., Kwak, S.: Promptstyler: Prompt-driven style generation for source-free domain generalization. In: {IEEE/CVF} International Conference on Computer Vision, {ICCV} 2023, Paris, France, October 1-6, 2023. pp. 15656--15666. {IEEE} (2023)

\bibitem{cimpoi14describing}
Cimpoi, M., Maji, S., Kokkinos, I., Mohamed, S., , Vedaldi, A.: Describing textures in the wild. In: Proceedings of the {IEEE} Conf. on Computer Vision and Pattern Recognition ({CVPR}) (2014)

\bibitem{deng2009imagenet}
Deng, J., Dong, W., Socher, R., Li, L.J., Li, K., Fei-Fei, L.: Imagenet: A large-scale hierarchical image database. In: 2009 IEEE conference on computer vision and pattern recognition. pp. 248--255. Ieee (2009)

\bibitem{dong2023maskclip}
Dong, X., Bao, J., Zheng, Y., Zhang, T., Chen, D., Yang, H., Zeng, M., Zhang, W., Yuan, L., Chen, D., et~al.: Maskclip: Masked self-distillation advances contrastive language-image pretraining. In: Proceedings of the IEEE/CVF Conference on Computer Vision and Pattern Recognition. pp. 10995--11005 (2023)

\bibitem{dosovitskiy2020image}
Dosovitskiy, A., Beyer, L., Kolesnikov, A., Weissenborn, D., Zhai, X., Unterthiner, T., Dehghani, M., Minderer, M., Heigold, G., Gelly, S., et~al.: An image is worth 16x16 words: Transformers for image recognition at scale. arXiv preprint arXiv:2010.11929  (2020)

\bibitem{FeiFei2004LearningGV}
Fei-Fei, L., Fergus, R., Perona, P.: Learning generative visual models from few training examples: An incremental bayesian approach tested on 101 object categories. Computer Vision and Pattern Recognition Workshop  (2004)

\bibitem{gal2016dropout}
Gal, Y., Ghahramani, Z.: Dropout as a bayesian approximation: Representing model uncertainty in deep learning. In: International Conference on Machine Learning. pp. 1050--1059. PMLR (2016)

\bibitem{gao2024clipadapter}
Gao, P., Geng, S., Zhang, R., Ma, T., Fang, R., Zhang, Y., Li, H., Qiao, Y.: Clip-adapter: Better vision-language models with feature adapters. International Journal of Computer Vision  \textbf{132}(2),  581--595 (2024)

\bibitem{gondal2024domain}
Gondal, M.W., Gast, J., Ruiz, I.A., Droste, R., Macri, T., Kumar, S., Staudigl, L.: Domain aligned clip for few-shot classification. In: Proceedings of the IEEE/CVF Winter Conference on Applications of Computer Vision. pp. 5721--5730 (2024)

\bibitem{GoyalCVPR23}
Goyal, S., Kumar, A., Garg, S., Kolter, Z., Raghunathan, A.: Finetune like you pretrain: Improved finetuning of zero-shot vision models. In: {IEEE/CVF} Conference on Computer Vision and Pattern Recognition, {CVPR} 2023, Vancouver, BC, Canada, June 17-24, 2023. pp. 19338--19347. {IEEE} (2023)

\bibitem{he2015deep}
He, K., Zhang, X., Ren, S., Sun, J.: Deep residual learning for image recognition. arxiv e-prints. arXiv preprint arXiv:1512.03385  \textbf{10} (2015)

\bibitem{helber2017eurosat}
Helber, P., Bischke, B., Dengel, A., Borth, D.: Eurosat: A novel dataset and deep learning benchmark for land use and land cover classification (2017)

\bibitem{hendrycks2021many}
Hendrycks, D., Basart, S., Mu, N., Kadavath, S., Wang, F., Dorundo, E., Desai, R., Zhu, T., Parajuli, S., Guo, M., Song, D., Steinhardt, J., Gilmer, J.: The many faces of robustness: A critical analysis of out-of-distribution generalization. ICCV  (2021)

\bibitem{hendrycks2016baseline}
Hendrycks, D., Gimpel, K.: A baseline for detecting misclassified and out-of-distribution examples in neural networks. arXiv preprint arXiv:1610.02136  (2016)

\bibitem{hendrycks2018deep}
Hendrycks, D., Mazeika, M., Dietterich, T.: Deep anomaly detection with outlier exposure. arXiv preprint arXiv:1812.04606  (2018)

\bibitem{hendrycks2021nae}
Hendrycks, D., Zhao, K., Basart, S., Steinhardt, J., Song, D.: Natural adversarial examples. CVPR  (2021)

\bibitem{jia2021scaling}
Jia, C., Yang, Y., Xia, Y., Chen, Y.T., Parekh, Z., Pham, H., Le, Q., Sung, Y.H., Li, Z., Duerig, T.: Scaling up visual and vision-language representation learning with noisy text supervision. In: International conference on machine learning. pp. 4904--4916. PMLR (2021)

\bibitem{khattak2023maple}
Khattak, M.U., Rasheed, H., Maaz, M., Khan, S., Khan, F.S.: Maple: Multi-modal prompt learning. In: Proceedings of the IEEE/CVF Conference on Computer Vision and Pattern Recognition. pp. 19113--19122 (2023)

\bibitem{khattak2023self}
Khattak, M.U., Wasim, S.T., Naseer, M., Khan, S., Yang, M.H., Khan, F.S.: Self-regulating prompts: Foundational model adaptation without forgetting. In: Proceedings of the IEEE/CVF International Conference on Computer Vision. pp. 15190--15200 (2023)

\bibitem{KrauseStarkDengFei-Fei_3DRR2013}
Krause, J., Stark, M., Deng, J., Fei-Fei, L.: 3d object representations for fine-grained categorization. In: 4th International IEEE Workshop on 3D Representation and Recognition (3dRR-13). Sydney, Australia (2013)

\bibitem{lafon2023hybrid}
Lafon, M., Ramzi, E., Rambour, C., Thome, N.: Hybrid energy based model in the feature space for out-of-distribution detection. In: International Conference on Machine Learning. pp. 18250--18268. PMLR (2023)

\bibitem{lee2018simple}
Lee, K., Lee, K., Lee, H., Shin, J.: A simple unified framework for detecting out-of-distribution samples and adversarial attacks. Advances in neural information processing systems  \textbf{31} (2018)

\bibitem{lu2022prompt}
Lu, Y., Liu, J., Zhang, Y., Liu, Y., Tian, X.: Prompt distribution learning. In: Proceedings of the IEEE/CVF Conference on Computer Vision and Pattern Recognition. pp. 5206--5215 (2022)

\bibitem{maji13fine-grained}
Maji, S., Kannala, J., Rahtu, E., Blaschko, M., Vedaldi, A.: Fine-grained visual classification of aircraft. Tech. rep. (2013)

\bibitem{ming2022delving}
Ming, Y., Cai, Z., Gu, J., Sun, Y., Li, W., Li, Y.: Delving into out-of-distribution detection with vision-language representations. Advances in Neural Information Processing Systems  \textbf{35},  35087--35102 (2022)

\bibitem{Miyai24}
Miyai, A., Yu, Q., Irie, G., Aizawa, K.: Locoop: Few-shot out-of-distribution detection via prompt learning. NeurIPS  \textbf{36} (2023)

\bibitem{Miyai23}
Miyai, A., Yu, Q., Irie, G., Aizawa, K.: Zero-shot in-distribution detection in multi-object settings using vision-language foundation models. CoRR  (2023)

\bibitem{Nie24}
Nie, J., Zhang, Y., Fang, Z., Liu, T., Han, B., Tian, X.: Out-of-distribution detection with negative prompts. In: The Twelfth International Conference on Learning Representations (2024)

\bibitem{Nilsback08}
Nilsback, M.E., Zisserman, A.: Automated flower classification over a large number of classes. In: Proceedings of the Indian Conference on Computer Vision, Graphics and Image Processing (Dec 2008)

\bibitem{parisot2023learning}
Parisot, S., Yang, Y., McDonagh, S.: Learning to name classes for vision and language models. In: Proceedings of the IEEE/CVF Conference on Computer Vision and Pattern Recognition. pp. 23477--23486 (2023)

\bibitem{parkhi12a}
Parkhi, O.M., Vedaldi, A., Zisserman, A., Jawahar, C.V.: Cats and dogs. In: IEEE Conference on Computer Vision and Pattern Recognition (2012)

\bibitem{Radford21}
Radford, A., Kim, J.W., Hallacy, C., Ramesh, A., Goh, G., Agarwal, S., Sastry, G., Askell, A., Mishkin, P., Clark, J., et~al.: Learning transferable visual models from natural language supervision. In: International conference on machine learning. pp. 8748--8763. PMLR (2021)

\bibitem{recht2019imagenet}
Recht, B., Roelofs, R., Schmidt, L., Shankar, V.: Do imagenet classifiers generalize to imagenet? In: International conference on machine learning. pp. 5389--5400. PMLR (2019)

\bibitem{SehwagCM21}
Sehwag, V., Chiang, M., Mittal, P.: {SSD:} {A} unified framework for self-supervised outlier detection. In: 9th International Conference on Learning Representations, {ICLR} 2021, Virtual Event, Austria, May 3-7, 2021 (2021)

\bibitem{ShuICML23}
Shu, Y., Guo, X., Wu, J., Wang, X., Wang, J., Long, M.: Clipood: Generalizing {CLIP} to out-of-distributions. In: ICML (2023)

\bibitem{soomro2012ucf101}
Soomro, K., Zamir, A.R., Shah, M.: Ucf101: A dataset of 101 human actions classes from videos in the wild. arXiv preprint arXiv:1212.0402  (2012)

\bibitem{srivastava2014dropout}
Srivastava, N., Hinton, G., Krizhevsky, A., Sutskever, I., Salakhutdinov, R.: Dropout: a simple way to prevent neural networks from overfitting. The journal of machine learning research  \textbf{15}(1),  1929--1958 (2014)

\bibitem{sun2022dualcoop}
Sun, X., Hu, P., Saenko, K.: Dualcoop: Fast adaptation to multi-label recognition with limited annotations. Advances in Neural Information Processing Systems  \textbf{35},  30569--30582 (2022)

\bibitem{sun2022out}
Sun, Y., Ming, Y., Zhu, X., Li, Y.: Out-of-distribution detection with deep nearest neighbors. In: International Conference on Machine Learning. pp. 20827--20840. PMLR (2022)

\bibitem{van2018inaturalist}
Van~Horn, G., Mac~Aodha, O., Song, Y., Cui, Y., Sun, C., Shepard, A., Adam, H., Perona, P., Belongie, S.: The inaturalist species classification and detection dataset. In: Proceedings of the IEEE conference on computer vision and pattern recognition. pp. 8769--8778 (2018)

\bibitem{villani2009optimal}
Villani, C., et~al.: Optimal transport: old and new, vol.~338. Springer (2009)

\bibitem{wang2022learning}
Wang, F., Li, M., Lin, X., Lv, H., Schwing, A., Ji, H.: Learning to decompose visual features with latent textual prompts. In: The Eleventh International Conference on Learning Representations (2022)

\bibitem{wang2019learning}
Wang, H., Ge, S., Lipton, Z., Xing, E.P.: Learning robust global representations by penalizing local predictive power. In: Advances in Neural Information Processing Systems. pp. 10506--10518 (2019)

\bibitem{Xiao:2010}
{Xiao}, J., {Hays}, J., {Ehinger}, K.A., {Oliva}, A., {Torralba}, A.: Sun database: Large-scale scene recognition from abbey to zoo. In: CVPR (2010)

\bibitem{ZhangECCV22}
Zhang, R., Zhang, W., Fang, R., Gao, P., Li, K., Dai, J., Qiao, Y., Li, H.: Tip-adapter: Training-free adaption of {CLIP} for few-shot classification. In: ECCV. pp. 493--510 (2022)

\bibitem{zhou2017places}
Zhou, B., Lapedriza, A., Khosla, A., Oliva, A., Torralba, A.: Places: A 10 million image database for scene recognition. IEEE Transactions on Pattern Analysis and Machine Intelligence  (2017)

\bibitem{ZhouLD22}
Zhou, C., Loy, C.C., Dai, B.: Extract free dense labels from {CLIP}. In: Avidan, S., Brostow, G.J., Ciss{\'{e}}, M., Farinella, G.M., Hassner, T. (eds.) Computer Vision - {ECCV} 2022 - 17th European Conference, Tel Aviv, Israel. Lecture Notes in Computer Science, vol. 13688, pp. 696--712. Springer (2022)

\bibitem{zhou2022conditional}
Zhou, K., Yang, J., Loy, C.C., Liu, Z.: Conditional prompt learning for vision-language models. In: Proceedings of the IEEE/CVF Conference on Computer Vision and Pattern Recognition. pp. 16816--16825 (2022)

\bibitem{zhou2022learning}
Zhou, K., Yang, J., Loy, C.C., Liu, Z.: Learning to prompt for vision-language models. International Journal of Computer Vision  \textbf{130}(9),  2337--2348 (2022)

\end{thebibliography}

\clearpage
\setcounter{section}{0}
\renewcommand\thesection{\Alph{section}}

\section{Additional details on method.}

In this section, we give additional details about \ours. In \cref{supp:clip_local}, we describe how the local features are extracted from CLIP's vision encoder, for both ResNet and ViT architectures. In \cref{supp:inference}, we describe the inference procedure in \ours as well as the GL-MCM score \cite{Miyai23} which we use for OOD detection. Finally, we discuss in \cref{supp:diversity_loss} the use of an additional explicit diversity loss to train \ours.

\subsection{CLIP's local visual features.}\label{supp:clip_local}

To obtain the visual local features from CLIP we follow previous works \cite{ZhouLD22, sun2022dualcoop, Miyai23}, which we describe in the following.

\noindent\textbf{ViT backbone.} When the vision encoder is a ViT, the output of the vision encoder is composed of the class token embedding, $\bm{z}_{\text{cls}}$, and a set of $L$ local features $\mathcal{Z}_l = (\bm{z}^l_1, ..., \bm{z}^l_L)$. The global visual representation used in CLIP is the class token embedding, \ie  $\bm{z}_g = \bm{z}_{\text{cls}}$, however the local features after the last transformer block are of low quality as only the class token receives a supervision signal during training. Hence, prior studies \cite{ZhouLD22, sun2022dualcoop, Miyai23} have recommended utilizing visual local features from the penultimate transformer block and forward them through the last transformer block without using the self attention mechanism.

Specifically, we have $\forall i \in \{1, ...,L\}$:
\begin{align*}
    \begin{split}
    \bm{z}^l_i ~&=~ \bm{z}^l_i + v(\bm{z}^l_i) ~+~ f(\bm{z}^l_i + v(\bm{z}^l_i)),\\
    \end{split}
\end{align*}
where $v(\cdot)$ denotes the linear projection used to compute the values in the self-attention module and $f(\cdot)$ is the feed-forward network of the last transformer block.\\

\noindent\textbf{ResNet backbone.} When the vision encoder is a ResNet the vision encoder outputs a feature map containing $L$ local patches $\mathcal{Z}_l = (\bm{z}^l_1, ..., \bm{z}^l_L)$. Then, the global visual feature, $\bm{z}_g$, is obtained using a self-attention pooling module: 
\begin{align*}
    \begin{split}
     \bm{z}_g &= \sum_{i} \text{softmax}(\frac{q(\overline{\bm{z}^l})~k(\bm{z}^l_i)^T}{\sqrt{d}})\cdot v(\bm{z}^l_i), \\
    \end{split}
\end{align*}
where $d$ is the feature dimension, $\overline{\bm{z}^l}=\frac{1}{L}\sum_{i=1}^L \bm{z}^l_i$ is the average-pooled feature used as unique query, and $q(\cdot)$, $k(\cdot)$, $v(\cdot)$ denote the query, key and value projections, respectively. To obtain useful visual local features, it is then sufficient to use the values of the local features without the attention mechanism, \ie $\bm{z}^l_i = v(\bm{z}^l_i)$.

\subsection{Details on \ours's inference.}\label{supp:inference}

In this section, we give more details on our inference procedure. As described in Sec. \textcolor{red}{3.2}, \ours is trained by summing the global and multiscale losses, associated to global and local prompts. Therefore, we naturally adopt an ``ensembling-style'' inference strategy by averaging the similarities obtained with each prompt to obtain a final similarity, $\text{sim}(\bm{z}, ~\bm{t}_c)$, for each class $y_c$.

Specifically, writing $\bm{z} = [\bm{z}_g,~ \mathcal{Z}_l]$, we compute:
\begin{align*}
     \text{sim}(\bm{z}, ~\bm{t}_c) &= \frac{1}{n} \sum_{i=1}^n \left<\bm{z}_g, ~\bm{t}_c(\bm{p^g}_i)\right> ~+~ \frac{1}{m} \sum_{j=1}^m \text{sim}_{\text{top-$k$}}(\mathcal{Z}_l, ~\bm{t}_c(\bm{p}^{l}_j)),
\end{align*}
where $\text{sim}_{\text{top-$k$}}(\mathcal{Z}_l, ~\bm{t}_c(\bm{p}^{l}_j))$ is defined in Eq. (\textcolor{red}{2}) of the main paper. Then with this final similarity computed, we use Eq. (\textcolor{red}{1}) of the main paper to compute the probability for class $y_c$.\\

To perform out-of-distribution detection with \ours we use the GL-MCM score \cite{Miyai23} which rely on both global and local information. The idea behind the MCM score \cite{ming2022delving} and the GL-MCM score \cite{Miyai23} is to perform a maximum concept matching, which is a natural extension of the maximum class probability (MCP) score \cite{hendrycks2016baseline} which is widely used baseline within the OOD community \cite{lee2018simple, SehwagCM21, sun2022out, lafon2023hybrid}.\\

Formally, the GL-MCM score is expressed as:
\begin{center}
    S\textsubscript{GL-MCM}~=~ S\textsubscript{G-MCM}  ~+~ S\textsubscript{L-MCM}
\end{center}
where
\begin{align*}
 S\textsubscript{G-MCM}~=~ \max_{c} ~ \frac{\exp( \frac{1}{n} \sum_{i=1}^n \left<\bm{z}_g, ~\bm{t}_c(\bm{p^g}_i)\right> ~/~ \tau)}{\sum_{c'} \exp (  \frac{1}{n} \sum_{i=1}^n \left<\bm{z}_g, ~\bm{t}_{c'}(\bm{p^g}_i)\right> ~/~ \tau)}, \\[12pt]
 S\textsubscript{L-MCM} ~= \max_{c, ~i} ~\frac{\exp( \frac{1}{m} \sum_{j=1}^m \left<\bm{z}^l_i, ~\bm{t}_c(\bm{p}^{l}_j)\right> ~/~ \tau)}{\sum_{c'} \exp ( \frac{1}{m} \sum_{j=1}^m \left<\bm{z}^l_i, ~\bm{t}_{c'}(\bm{p}^{l}_j)\right> ~/~ \tau)}.
\end{align*}

\vspace{1em}
\subsection{Diversity loss.}\label{supp:diversity_loss}
Previous works on prompt ensembling have explored the use of an explicit loss term encouraging the semantic orthogonality between prompts to increase their diversity \cite{lu2022prompt, ChoICCV23}. This loss is expressed as:
\begin{equation*}
    \mathcal{L}_{\text{div.}}(\mathcal{P}) = \frac{1}{N\cdot(N-1)} \sum_{i=1}^N \sum_{j=i+1}^N  |\left<\bm{t}_i, \bm{t}_j\right>|,
\end{equation*}

where $\mathcal{P}$ is a set of $N$ prompts and $\forall i \in \{1,\cdots, N\},~ \bm{t}_i$ are the textual representations of the prompts without incorporating class names. The strength of the diversity loss is controlled with a hyper-parameter $\lambda_\text{div.}$.\\

\begin{figure}[h]
        \centering
        \includegraphics[width=0.5\linewidth]{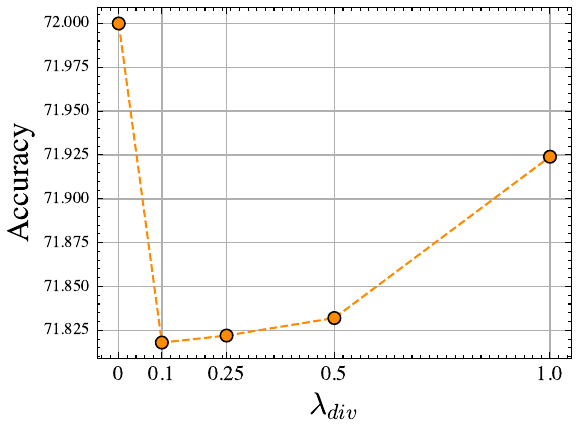}
    \caption{Impact of $\lambda_\text{div.}$.}
    \label{fig:abaltion_lambda_div}
\end{figure}

We have experimented optimizing \ours with the following loss: $\mathcal{L}_{\text{total}}(\mathcal{P},~\bm{\theta}) + \lambda_{\text{div.}} \cdot \mathcal{L}_{\text{div.}}(\mathcal{P})$. In \cref{fig:abaltion_lambda_div}, we show that training \ours with $\mathcal{L}_{\text{div}}$ does not improve top-1 accuracy, even when increasing $\lambda_{\text{div.}}$. As a result, we did not include $\mathcal{L}_{\text{div.}}$ in \ours, as it did not lead to significant improvement in either accuracy or robustness, while introducing an extra hyperparameter, $\lambda_{\text{div.}}$. 

\vspace{2em}

\section{Additional experimental results.}

In this section, we give additional experimental results of \ours. In \cref{sec:supp_few_shots} we conduct more results for few-shots settings experiments on the suite 11 of datasets. In \cref{sec:backbones} we give results of \ours when using a ResNet-50 CLIP backbone. In \cref{sec:sup_dg} and \cref{sec:sup_ood} we give the detailed results for the ImageNet-1k domain generalization and out-of-distribution detection benchmarks, respectively. In \cref{sec:other_few_shots_methods} we compare \ours to other few-shots learning methods. Finally, we show additional qualitative results in \cref{sec:sup_quali}.

\subsection*{Additional implementation details.} In this section, we give more implementation details of \ours. We show on \cref{tab:hyperparameters} the hyperparameters used to train \ours on ImageNet for the 16-shots setting. We use the same data augmentation than CoOp \cite{zhou2022learning}.

\begin{table}[h]
    \centering
    \caption{Hyperparameters to train \ours on ImageNet (16 shots) with ViT-B/16 backbone.}
    \label{tab:hyperparameters}
    \begin{tabularx}{0.6\textwidth}{YY}
        \toprule
        \textbf{Hyperparameters} & \textbf{Value} \\
        \midrule
        batch size & 128 \\
        learning rate & 0.002 \\
        lr-scheduler & CosineAnnealingLR \\
        epochs & 50 \\
        optimizer & SGD \\
        weight decay & 0.01 \\
        momentum & 0.9 \\
        local prompts & 4 \\
        global prompts & 4 \\
        tokens per prompt & 4 \\
        prompt init & ``A photo of a''\\
        \bottomrule
    \end{tabularx}
\end{table}

\subsection{Full few shot results.}\label{sec:supp_few_shots} 
\vspace{1em}
In this section, we give the detailed results for different few-shots settings. We report the top-1 accuracy of \ours on each dataset of the few-shot learning benchmark introduced in \cite{zhou2022learning}. We can see in \cref{tab:sup_average_few_shots} that \ours outperforms other prompt learning baselines for all shots on average on the suite of 11 datasets. Specifically, \ours consistently outperforms the second-best method PromptSRC by +0.5pt with 1-shot, +1.1pt with 2-shots, +0.8pt with 4-shots, +1.5pt with 8-shots and +1.6pt with 16-shots. All results for each of the 11 datasets are ploted on \cref{fig:low_shot_experiments}.

\begin{table}
    \centering
    \caption{Averaged few-shots results on the suite 11 datasets with ViT-B/16 backbone.}
    \label{tab:sup_average_few_shots}
    \begin{tabularx}{0.78\linewidth}{l YYYYYY}
        \toprule
         \textbf{Method}&  \small{0-shot}&  \small{1-shot}&  \small{2-shots}& \small{ 4-shots}&  \small{8-shots}& \small{16-shots}\\
         \midrule
         CLIP & 64.9 & - & - & - & - & - \\
         CoOp & - & 67.6  & 70.6 & 74.0 & 77.0 & 79.9\\
         MaPLe & - & 69.3 & 72.6 & 75.8 & 78.9 & 81.8 \\
         PLOT & - & 70.7 & 74.0 & 76.9 & 79.6 & 82.1 \\
         PromptSRC & - & 72.3& 75.3& 78.3& 80.7& 82.9 \\
        \rowcolor{mycornflowerblue} \textbf{\ours}  & - & \textbf{72.8} & \textbf{76.4} & \textbf{79.1} & \textbf{82.2} & \textbf{84.5} \\
         \bottomrule
    \end{tabularx}
\end{table}

\newcommand\SupPlotSize{0.325}

\begin{figure}
    \centering
    
    \begin{subfigure}[t]{\SupPlotSize\textwidth}
        \includegraphics[width=\textwidth]{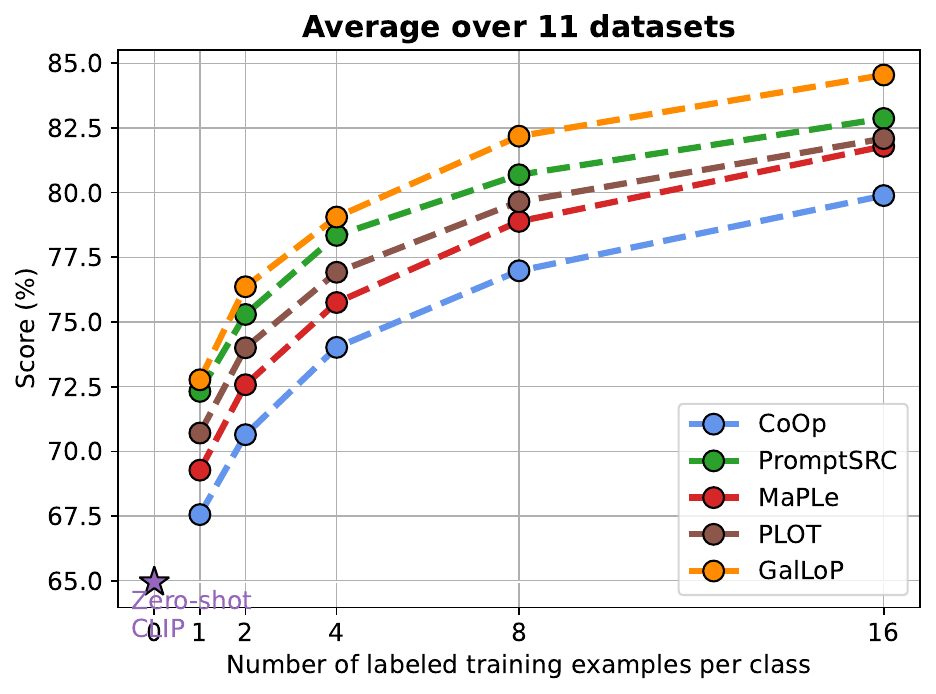}
        \label{fig:low_shot_average}
    \end{subfigure}
    \hfill
    \begin{subfigure}[t]{\SupPlotSize\textwidth}
        \includegraphics[width=\textwidth]{illustrations/low_shot_experiments/ImageNet.pdf}
        \label{fig:low_shot_imagenet}
    \end{subfigure}
    \hfill
    \begin{subfigure}[t]{\SupPlotSize\textwidth}
        \includegraphics[width=\textwidth]{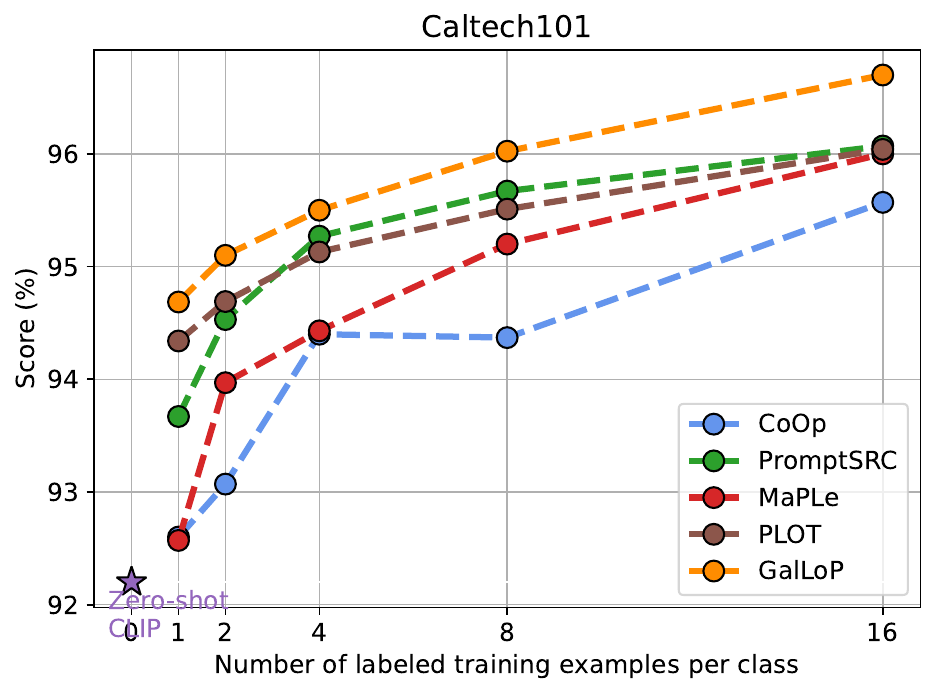}
        \label{fig:low_shot_caltech}
    \end{subfigure}
    
    \vspace{1em}
    
    \begin{subfigure}[t]{\SupPlotSize\textwidth}
        \includegraphics[width=\textwidth]{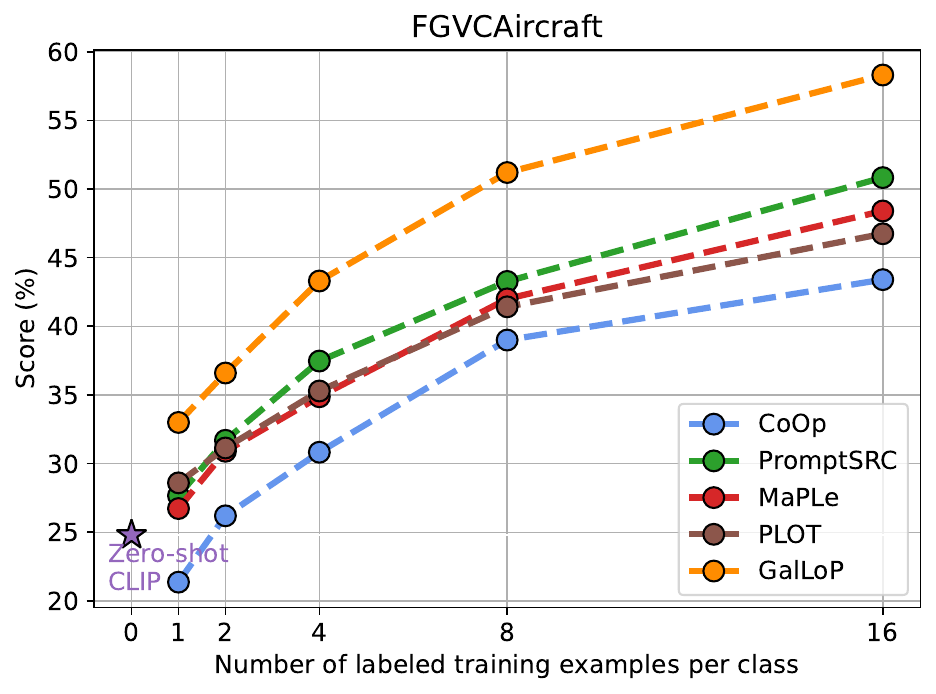}
        \label{fig:low_shot_aircraft}
    \end{subfigure}
    \hfill
    \begin{subfigure}[t]{\SupPlotSize\textwidth}
        \includegraphics[width=\textwidth]{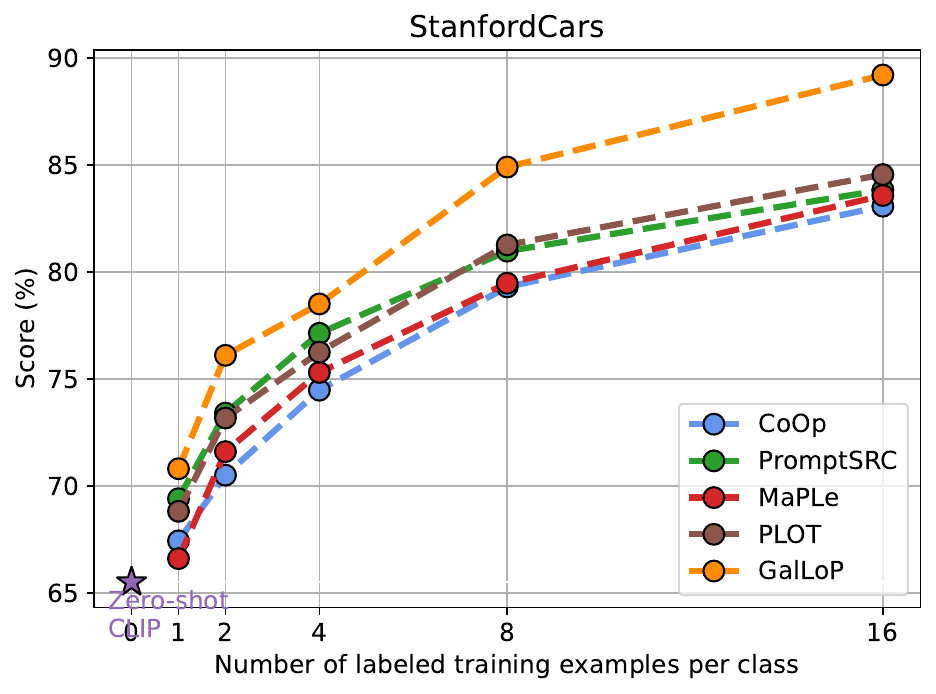}
        \label{fig:low_shot_cars}
    \end{subfigure}
    \hfill
    \begin{subfigure}[t]{\SupPlotSize\textwidth}
        \includegraphics[width=\textwidth]{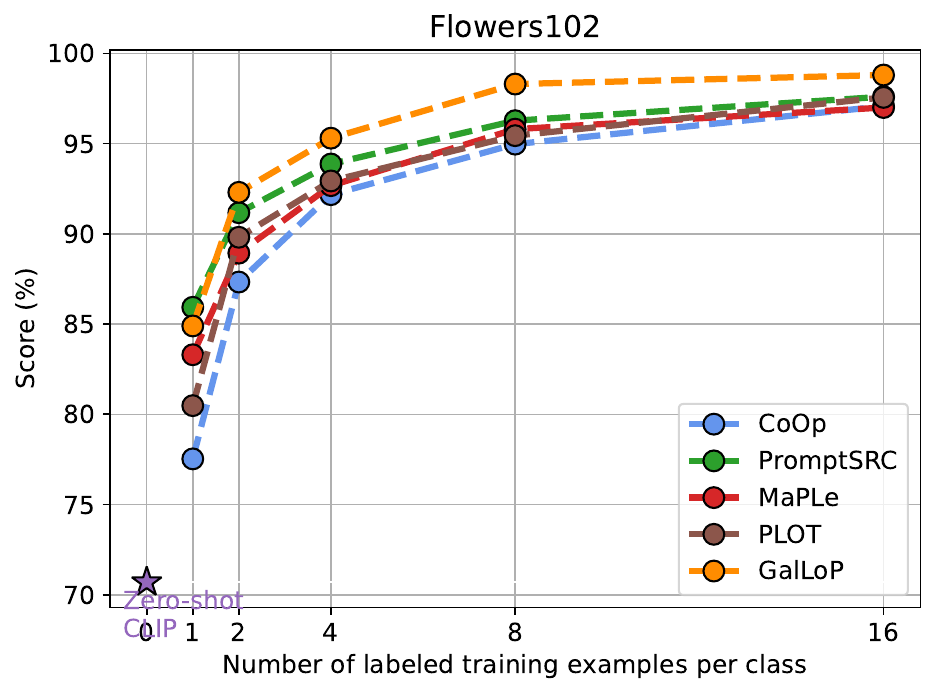}
        \label{fig:low_shot_flowers}
    \end{subfigure}

    \vspace{1em}
        
    \begin{subfigure}[t]{\SupPlotSize\textwidth}
        \includegraphics[width=\textwidth]{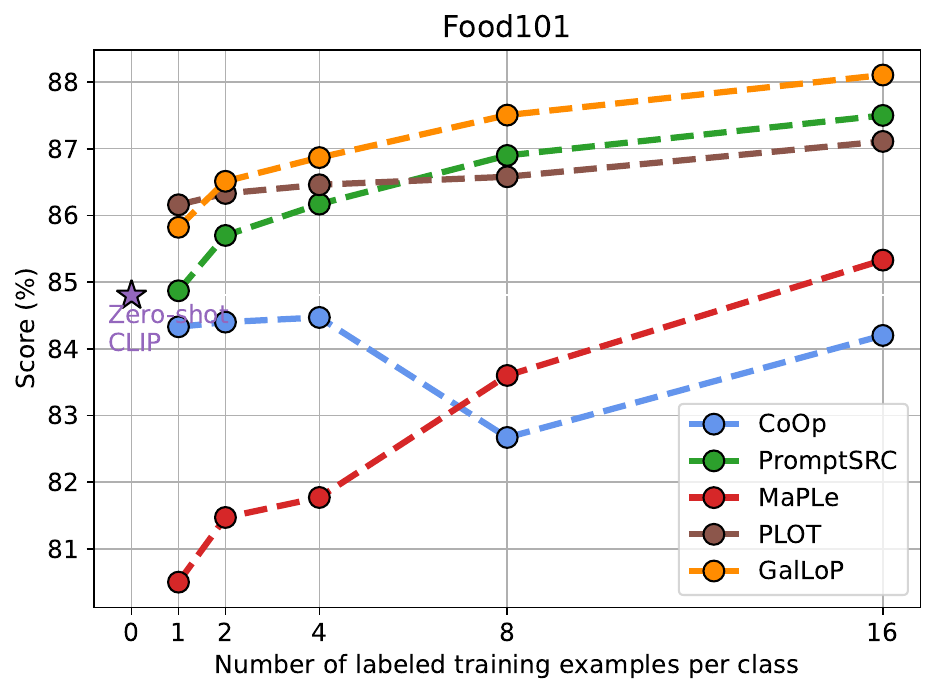}
        \label{fig:low_shot_food}
    \end{subfigure}
    \hfill
    \begin{subfigure}[t]{\SupPlotSize\textwidth}
        \includegraphics[width=\textwidth]{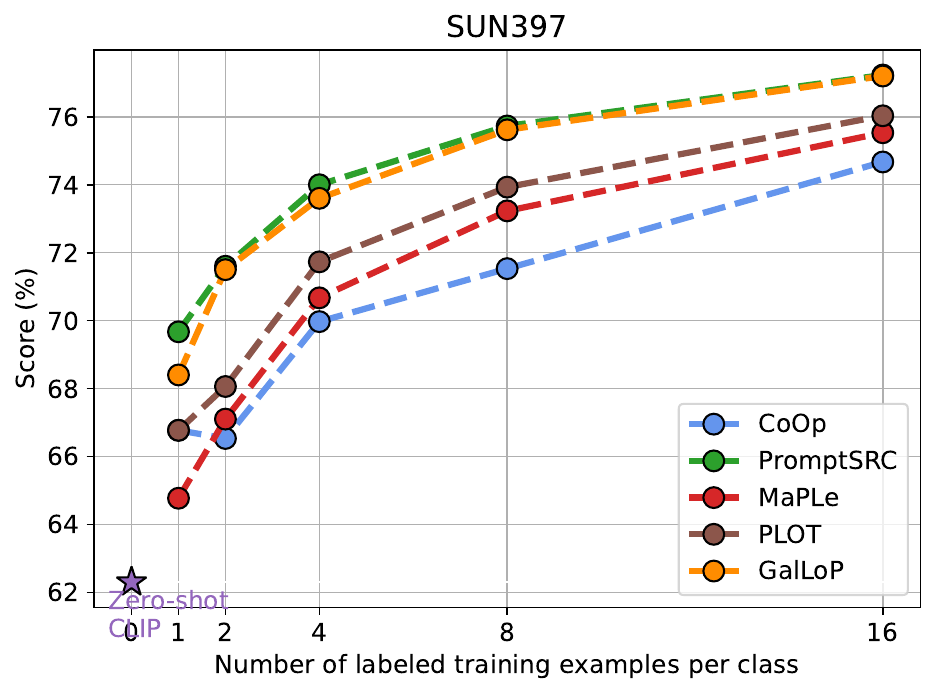}
        \label{fig:low_shot_sun}
    \end{subfigure}
    \hfill
    \begin{subfigure}[t]{\SupPlotSize\textwidth}
        \includegraphics[width=\textwidth]{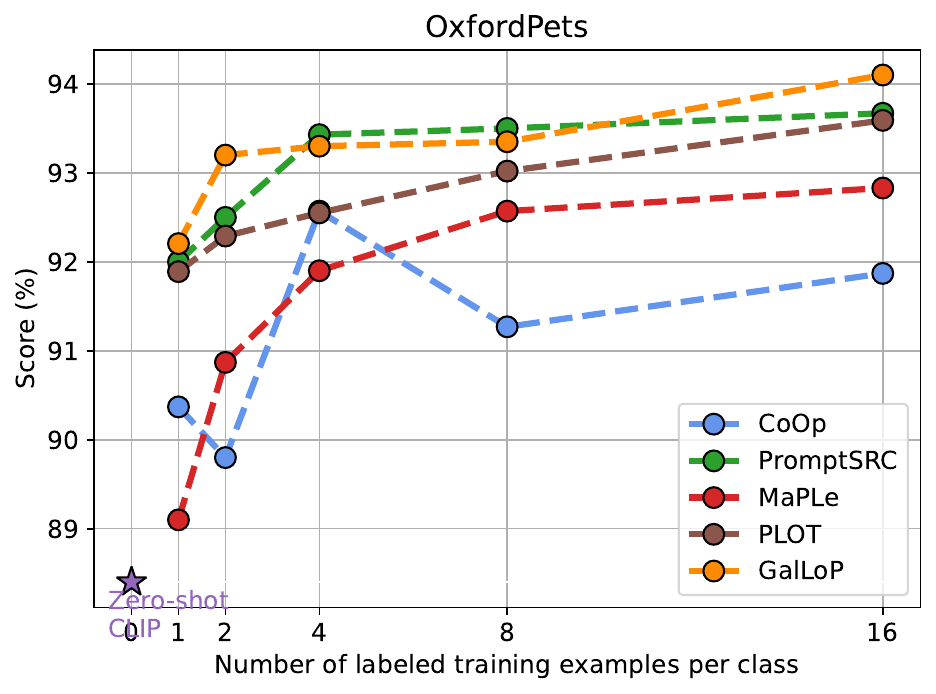}
        \label{fig:low_shot_pets}
    \end{subfigure}

    \vspace{1em}
    \begin{subfigure}[t]{\SupPlotSize\textwidth}
        \includegraphics[width=\textwidth]{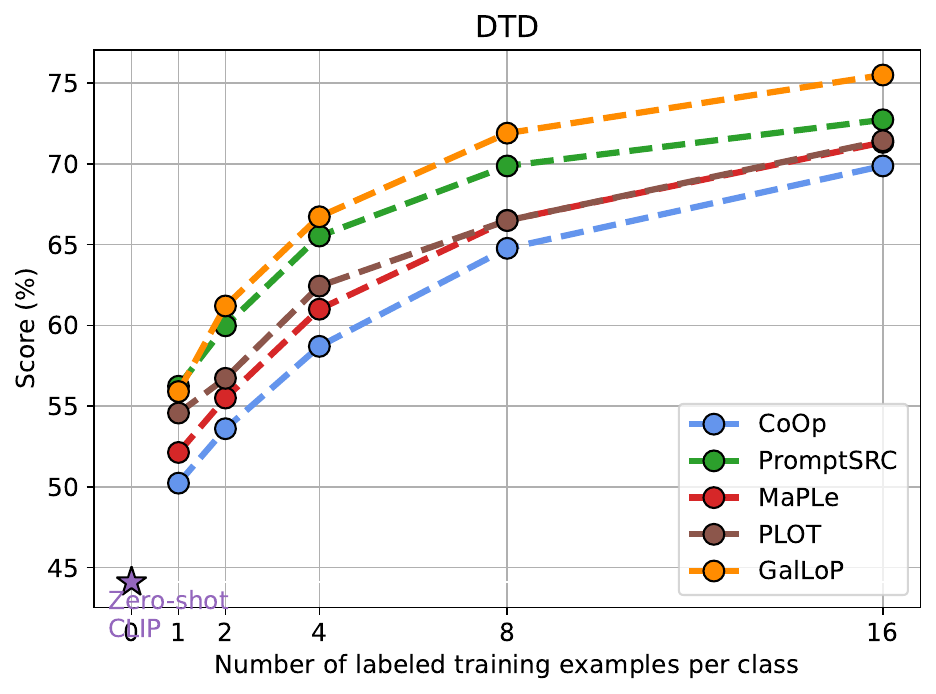}
        \label{fig:low_shot_dtd}
    \end{subfigure}
    \hfill
    \begin{subfigure}[t]{\SupPlotSize\textwidth}
        \includegraphics[width=\textwidth]{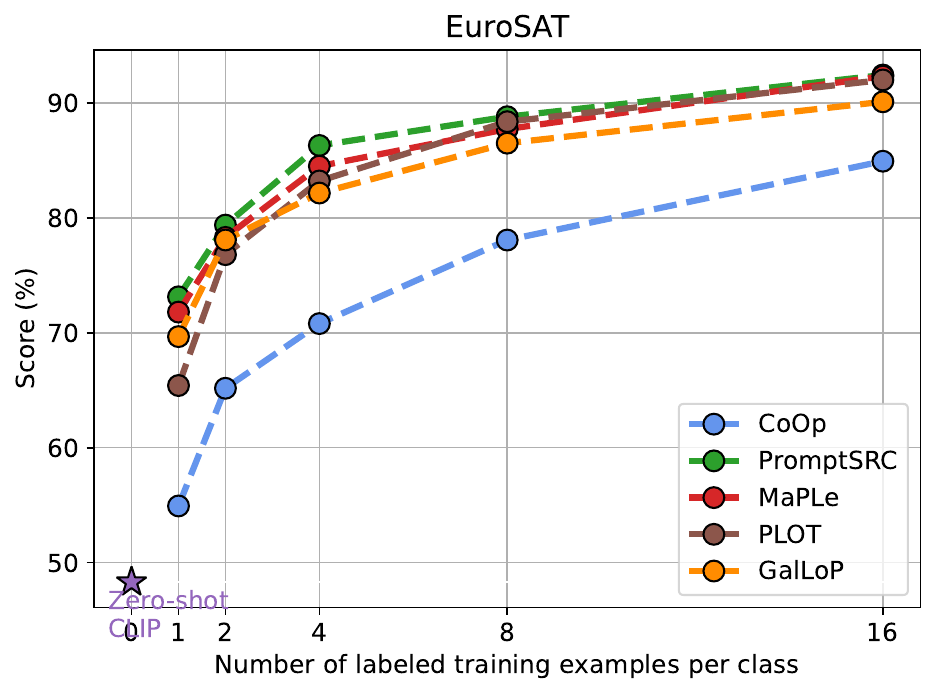}
        \label{fig:low_shot_eurosat}
    \end{subfigure}
    \hfill
    \begin{subfigure}[t]{\SupPlotSize\textwidth}
        \includegraphics[width=\textwidth]{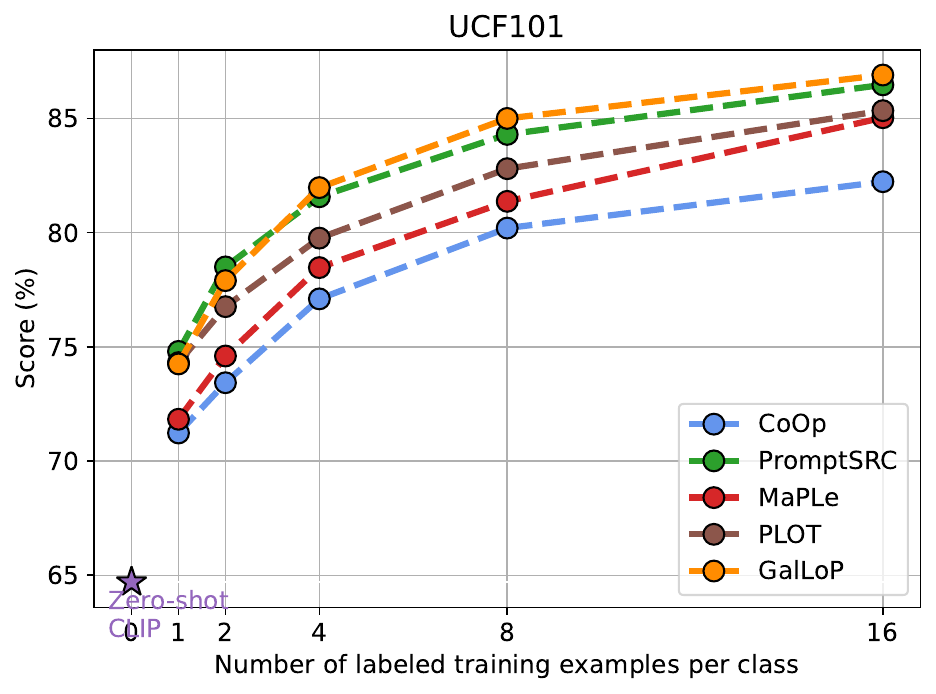}
        \label{fig:low_shot_ucf}
    \end{subfigure}
    \vspace{-1em}
    \caption{Few-shot learning results of \ours on the 11 datasets with the ViT-B/16 backbone.}
    \label{fig:low_shot_experiments}
\end{figure}

\subsection{Detailed results for ResNet-50.}\label{sec:backbones} 
In this section we give detailed results for \ours when trained using a ResNet-50 backbone. We can see in \cref{tab:top1_accuracy_resnet} that \ours outperforms other ResNet-50 compatible prompt learning methods on all the datasets except Food101. Specifically, \ours  achieves 77.3\% accuracy on average on the suite 11 of datasets, outperforming PLOT by +3.4pt and CoOp by +3.9pt. Note that the second-best method on ViT-B/16, PromptSRC~\cite{khattak2023self}, is not compatible with convolutional backbones such as the ResNet-50.

\begin{table}[ht]
    \setlength\tabcolsep{0.5pt}
    \caption{Top-1 accuracy with a Resnet-50 backbone in the 16-shots setting. Comparison of \ours to other prompt learning methods on the suite of 11 datasets.}
    \vspace{-1em}
    \label{tab:top1_accuracy_resnet}
    \centering
    \begin{tabularx}{\linewidth}{l YYYYYYYYYYYYY}
        \toprule

         Dataset & \rotatebox[origin=c]{70}{ImageNet~\cite{deng2009imagenet}} & \rotatebox[origin=c]{70}{Caltech101~\cite{FeiFei2004LearningGV}} & \rotatebox[origin=c]{70}{OxfordPets~\cite{parkhi12a}} & \rotatebox[origin=c]{70}{Cars~\cite{KrauseStarkDengFei-Fei_3DRR2013}} & \rotatebox[origin=c]{70}{Flowers102~\cite{Nilsback08}} & \rotatebox[origin=c]{70}{Food101~\cite{bossard14}} & \rotatebox[origin=c]{70}{Aircraft~\cite{maji13fine-grained}} & \rotatebox[origin=c]{70}{SUN397~\cite{Xiao:2010}} & \rotatebox[origin=c]{70}{DTD~\cite{cimpoi14describing}} & \rotatebox[origin=c]{70}{EuroSAT~\cite{helber2017eurosat}} & \rotatebox[origin=c]{70}{UCF101~\cite{soomro2012ucf101}} & \rotatebox[origin=c]{70}{Average} \\
         \midrule
         CLIP & 58.1 & 84.1 & 82.7 & 55.8 & 66.0 & 75.0 & 17.0 & 57.1 & 42.9 & 36.3 & 57.9 & 57.5 \\
         Linear Probe &  55.9 & 90.6 & 76.4 & 70.1 & 95.0 & 70.2 & \underline{36.4} & 67.2 & 64.0 & 82.8 & 73.7 & 71.1\\
         CoOp & \underline{63.0} & 91.8 & 87.0 & \underline{73.4} & 94.5 & 74.7 & {31.3} & 69.3 & 63.6 & \underline{83.5} & 75.7 & 73.4 \\
         Co-CoOp & 62.9 & 90.2 & 88.3 & 61.6 & 78.3 & \textbf{80.0} & 21.3 & 67.3 & 56.2 & 70.1 & 71.1  & 67.9\\
         PLOT & \underline{63.0} & \underline{92.2} & \underline{87.2} & 72.8 & \underline{94.8} & \underline{77.1} & {31.5} & \underline{70.0} & \underline{65.6} & 82.2 & \underline{77.3} & \underline{73.9} \\[2pt]
         \rowcolor{mycornflowerblue} \textbf{\ours} & \textbf{66.1} & \textbf{92.8} & \textbf{89.3} & \textbf{79.3} & \textbf{96.7} & {76.5} & \textbf{41.6} & \textbf{72.2} & \textbf{67.6} & \textbf{87.6} & \textbf{80.4} & \textbf{77.3} \\
         \bottomrule
    \end{tabularx}
\end{table}

\newpage
\subsection{\ours \vs other few-shots learning methods.}\label{sec:other_few_shots_methods}

In this section, we compare \ours to other type of few-shots learning methods. We compare \ours against the standard fine-tuning of all the parameters of CLIP's vision and text encoders as well as FLYP \cite{GoyalCVPR23}, which is a more recent version using the same contrastive objective as CLIP to fine-tune on downstream datasets. We also consider CLIP\textsubscript{OOD} \cite{ShuICML23}, which only trains the visual encoder. Furthermore, we also include adapters, \eg the recent CLIP-Adapter \cite{gao2024clipadapter}, which uses residual adapters on both the visual and textual representations. Finally, we compare against cached-based methods like Tip-Adapter / Tip-Adapter-F \cite{ZhangECCV22} and DAC-V / DAC-VT \cite{gondal2024domain}.

We show in \cref{fig:sup_cache_hp_count} the performance of \ours \vs the other few-shots learning methods in the 16-shots setting using a ViT-B/16 backbone. We can see that \ours obtains better top-1 accuracy than the recent fine-tuning method FLYP while it fine-tunes $\times 250$ more parameters than \ours. Furthermore, \ours outperforms the best cache-based method, DAC-VT, by +0.5pt, while having half the number of parameters. Also, when compared to DAC-V, which has the same number of parameters, \ours obtains +2.1pt in top-1 accuracy.

\begin{table}[ht]
    \setlength\tabcolsep{0.45pt}
    \centering
    \caption{Comparison of \ours \vs other few-shots learning  methods on ViT-B/16 in the 16-shots setting.}
    \label{fig:sup_cache_hp_count}
     \vspace{-1em}
        \begin{tabularx}{0.6\linewidth}{l YYY}
        \toprule
         & Top-1 & \# params \scriptsize{($\times$10$^6$)} \\
         \midrule
         \small{Zero-Shot CLIP} & 68.6 & 0 \\  
         Tip-Adapter \cite{ZhangECCV22} & 70.8 & 0 \\
         \midrule
         Full fine-tuning \cite{GoyalCVPR23} & 73.1 & 149.7\\
         CLIP\textsubscript{OOD} \cite{ShuICML23} &71.6& 86.7\\
         FLYP \cite{GoyalCVPR23} & \underline{74.9} & 149.7\\
         \midrule
         CLIP-Adapter \cite{gao2024clipadapter}  & 71.1 &  0.2 \\         
         \small{Tip-Adapter-F}  \cite{ZhangECCV22} & 73.7 & 16.4 \\
         DAC-V \cite{gondal2024domain} & 73.0 & 0.6\\
         DAC-VT \cite{gondal2024domain} & {74.6} &  1.1 \\
         \rowcolor{mycornflowerblue} \textbf{\ours} & \textbf{75.1} & 0.6 \\
         \bottomrule
        \end{tabularx}
\end{table}

\subsection{Detailed domain generalization results.}\label{sec:sup_dg} In this section, we give the detailed results for the ImageNet domain generalization benchmark. We compare the performances of \ours with several prompt learning methods. Each method is trained on ImageNet with 16 shots per class and is evaluated on top-1 accuracy on four variants of ImageNet, \ie ImageNet-V2~\cite{recht2019imagenet}, ImageNet-Sketch~\cite{wang2019learning}, ImageNet-A~\cite{hendrycks2021nae} and ImageNet-R~\cite{hendrycks2021many}. We can see in \cref{tab:domain_generalization_vit} that \ours outperforms previous prompt learning methods on average on the four ImageNet variants with +0.6pt top1-accuracy \vs PromptSRC$^\diamond$. More specifically, we obtain better results on ImageNet-V2, with +1.8pt with respect to the second-best method, and comparable results to PromptSRC$^\diamond$ on ImageNet-Sketch and ImageNet-R.

\subsection{Detailed OOD detection results.}\label{sec:sup_ood} In this section, we give the detailed results of \ours for OOD detection. We use the OOD detection benchmark from \cite{ming2022delving} where ImageNet-1k is the in-distribution (ID) dataset, and iNaturalist~\cite{van2018inaturalist}, SUN~\cite{Xiao:2010}, Places~\cite{zhou2017places} and Textures~\cite{cimpoi14describing} are used as OOD datasets. We report the results using the FPR95$\downarrow$ and the AUC$\uparrow$ metrics, two standard metrics used by the OOD detection community. The FPR95 is the false positive rate, using a threshold corresponding that classifies 95\% of the ID images correctly. The AUC is the area under the receiver operating characteristic curve (ROC). We can see in \cref{tab:ood_detection_vit} that \ours obtains better averaged FPR95 results than other prompt learning methods with -1.4pt \vs LoCoOp while achieving 93.2 averaged AUC, the second-best result after LoCoOp (93.5 averaged AUC).

\begin{table}[ht]
    \setlength\tabcolsep{0.4pt}
    \caption{Domain generalization from ImageNet with ViT-B/16 backbone. Prompt learning methods are trained on ImageNet and evaluated on datasets with domain shifts. $^\dagger$results based on our re-implementation.}
    \label{tab:domain_generalization_vit}
    \centering
    \vspace{-1em}
    \begin{tabularx}{\linewidth}{l YYYYYY}
        \toprule
         \multirow{2}{*}{} & \multicolumn{1}{c}{Source} & \multicolumn{5}{c}{Target} \\
         \cmidrule(lr){2-2} \cmidrule(lr){3-7}
         & ImageNet & -V2~\cite{recht2019imagenet} & -S~\cite{wang2019learning} & -A~\cite{hendrycks2021nae} & -R~\cite{hendrycks2021many} & Avg. \\
         \midrule
         CLIP & 66.7 & 60.8 & 46.2 & 47.8 & 74.0 & 57.2 \\
         CoOp & 71.7 & 64.6 & 47.9 & 49.9 & 75.1 & 59.4 \\
         Co-CoOp & 71.0 & 64.1 & 48.8 & \underline{50.6} & 76.2 & 59.9 \\
         MaPLe & 70.7 & 64.1 & \underline{49.2} & \textbf{50.9} & 77.0 & 60.3 \\
         PLOT &  {72.6} & 64.9 & 46.8 & 48.0 & 73.9 & 58.4\\
         PromptSRC$^\diamond$ & 71.3 & {64.4} & \textbf{49.6} & \textbf{50.9} & \textbf{77.8} & \underline{60.7} \\
         PromptSRC$^\triangleright$ & \underline{73.2} & \underline{65.7} & 49.1 & 47.6 & 76.9 & 59.8 \\
         LoCoOp$^\dagger$ & 71.5 & 64.7 & 47.4 & 49.8 & 75.0 & 57.5\\
         ProDA$^\dagger$ & 71.9 & 64.5 & 48.6 & \underline{50.7} & 76.3 & 60.0\\
         \rowcolor{mycornflowerblue}  \textbf{\ours} & \textbf{75.1} & \textbf{67.5} & \textbf{49.5} & 50.3 & \textbf{77.8} & \textbf{61.3} \\
         \bottomrule
        \end{tabularx}
\end{table}

\begin{table}[ht]
    \setlength\tabcolsep{0.6pt}
    \caption{OOD detection with ViT-B/16 as backbone. CoOp and LoCoOp results reported from \cite{Miyai24}. CoCoOp and LSN results are reported from \cite{Nie24}. $^\dagger$ denotes results based on our re-implementation. For PLOT, we use their released checkpoint and evaluate its OOD detection results ourselves.}
    \label{tab:ood_detection_vit}
    \centering
    \vspace{-1em}
    \begin{tabularx}{\linewidth}{l YYYYYYYYYY Y}
        \toprule
         \multirow{2}{*}{} & \multicolumn{2}{c}{iNat~\cite{van2018inaturalist}} & \multicolumn{2}{c}{SUN~\cite{Xiao:2010}} & \multicolumn{2}{c}{Places~\cite{zhou2017places}} & \multicolumn{2}{c}{Textures~\cite{cimpoi14describing}} & \multicolumn{2}{c}{\textbf{Average}} & \multirow{2}{*}{Top-1} \\
         \cmidrule(lr){2-3} \cmidrule(lr){4-5} \cmidrule(lr){6-7} \cmidrule(lr){8-9} \cmidrule(lr){10-11}
         & {\tiny FPR95$\downarrow$} & {\tiny AUC$\uparrow$} & {\tiny FPR95$\downarrow$} & {\tiny AUC$\uparrow$} & {\tiny FPR95$\downarrow$} & {\tiny AUC$\uparrow$} & {\tiny FPR95$\downarrow$} & {\tiny AUC$\uparrow$} & {\tiny FPR95$\downarrow$} & {\tiny AUC$\uparrow$}\\
         \midrule
         MCM & 30.9 & 94.6 & 37.7 & 92.6 & 44.8 & 89.8 & 57.9 & 86.1 & 42.8 & 90.8 & 66.7 \\
         GL-MCM & {15.2} & 96.7 & 30.4 & 93.1 & 38.9 & 89.9 & 57.9 & 83.6 & 35.5 & 90.8 & 66.7 \\
         \midrule
         PLOT & {15.9} & {96.6} & 33.7 & 92.8 & 38.2 & 91.0 & {39.2} & {90.2}&   31.8 & {92.7} & 72.6 \\
         PromptSRC$^\diamond$ & 28.8 & 93.9 &  35.9 & 92.6 &  42.4 & 90.0 &  46.9 & 88.9 & 38.5 & 91.4 & 71.3 \\
         PromptSRC$^\triangleright $ & 20.6 & 95.7&  30.1 & {93.7}&  38.0 & {91.1}&  46.0 & 89.0 &   33.7 & 92.4 & \underline{73.2} \\
         ProDA$^\dagger$ & 32.4 & 93.2 & 35.7 & 92.4 & 42.6 & 90.0 & 46.2 & 89.3 & 39.2 & 91.2 & 71.9 \\
         CoOp\textsubscript{MCM} & 28.0 & 94.4 & 37.0 & 92.3 & 43.0 & 89.7 & 39.3 & \textbf{91.2} & 36.8 & 91.9 & 71.7 \\
         CoOp\textsubscript{GL} & \underline{14.6} & 96.6 & 28.5 & 92.7 & 36.5 & 90.0 & 43.1 & 88.0 & 30.7 & 91.8 & 71.7 \\
         CoCoOp & 30.7 & 94.7 & 31.2 & 93.2& 38.8 & 90.6 & 53.8 & 87.9 &38.6 & 91.6 & 71.0 \\
         LoCoOp\textsubscript{MCM} & 23.1 & 95.5 & 32.7 & 93.4 & 39.9 & 90.6 & 40.2 & \textbf{91.3} & 34.0 & 92.7 & 71.5 \\
         LoCoOp\textsubscript{GL} & 16.1 & \underline{96.9} & \textbf{23.4} & \textbf{95.1} & \underline{32.9} & \textbf{92.0} & 42.3 & 90.2 & {28.7} & \textbf{93.5} & 71.5 \\
         LSN\textsubscript{+CoOp} &  23.5 & 95.5 & 29.8 & 93.5 & 36.4 & 90.9 &  \textbf{38.2} & 89.5& 32.0 & 92.3 & 72.9 \\
         LSN\textsubscript{+CoCoOp}&  21.6 & 95.8 & 26.3 & \underline{94.4} & 34.5 & \underline{91.3} & 38.5 & \underline{90.4}  & 30.2 & 93.0 & 71.9 \\
         \rowcolor{mycornflowerblue}  \textbf{\ours} &  \textbf{13.7} & \textbf{97.1} & \underline{24.9} & {94.0} & \textbf{32.5} & \underline{91.3} & \underline{38.4} & \underline{90.4} & \textbf{27.3} & \underline{93.2} & \textbf{75.1} \\
         \bottomrule
    \end{tabularx}
\end{table}

\newpage
\subsection{Additional qualitative results.}\label{sec:sup_quali} 
\begin{figure}[h]
\centering
\begin{tabular}{c}
 \includegraphics[width=0.85\linewidth]{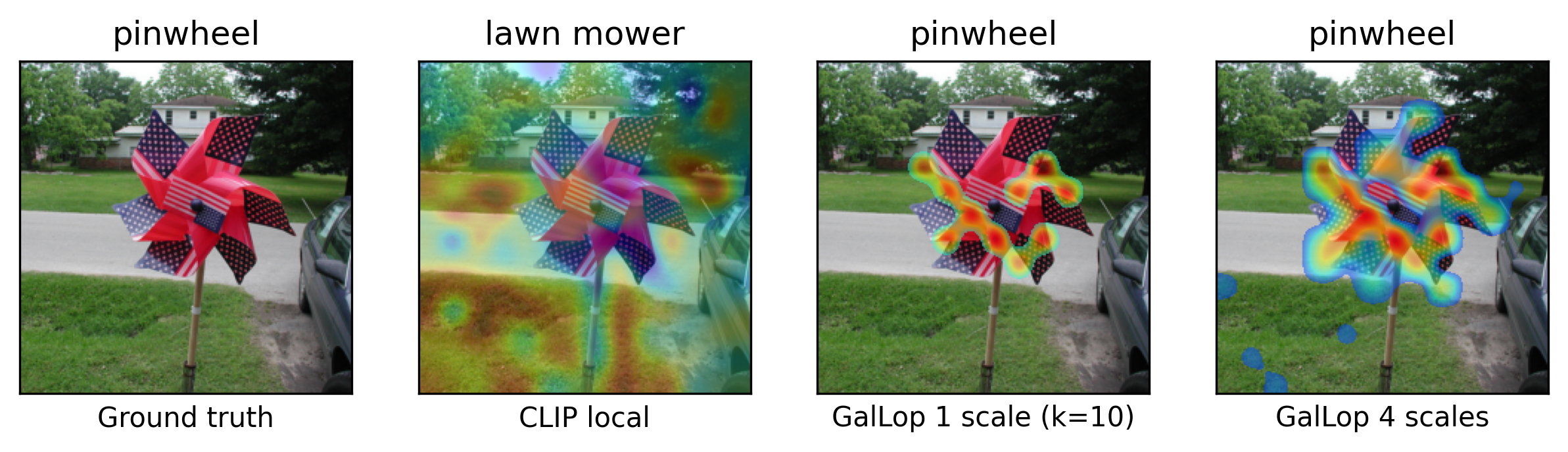} \\
 \includegraphics[width=0.85\linewidth]{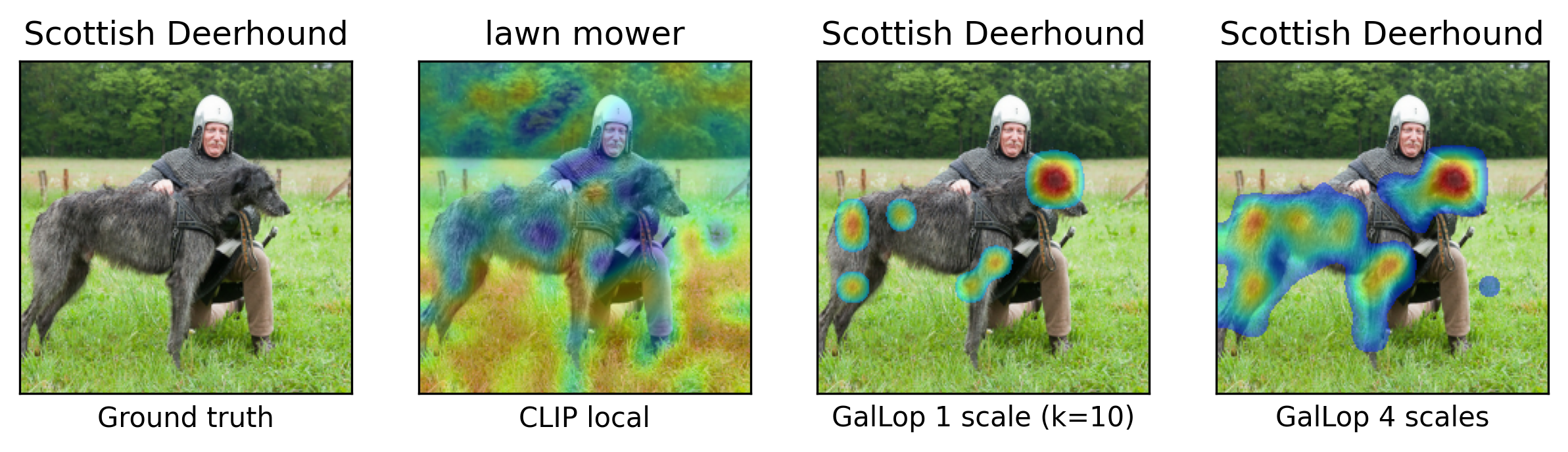} \\
 \includegraphics[width=0.85\linewidth]{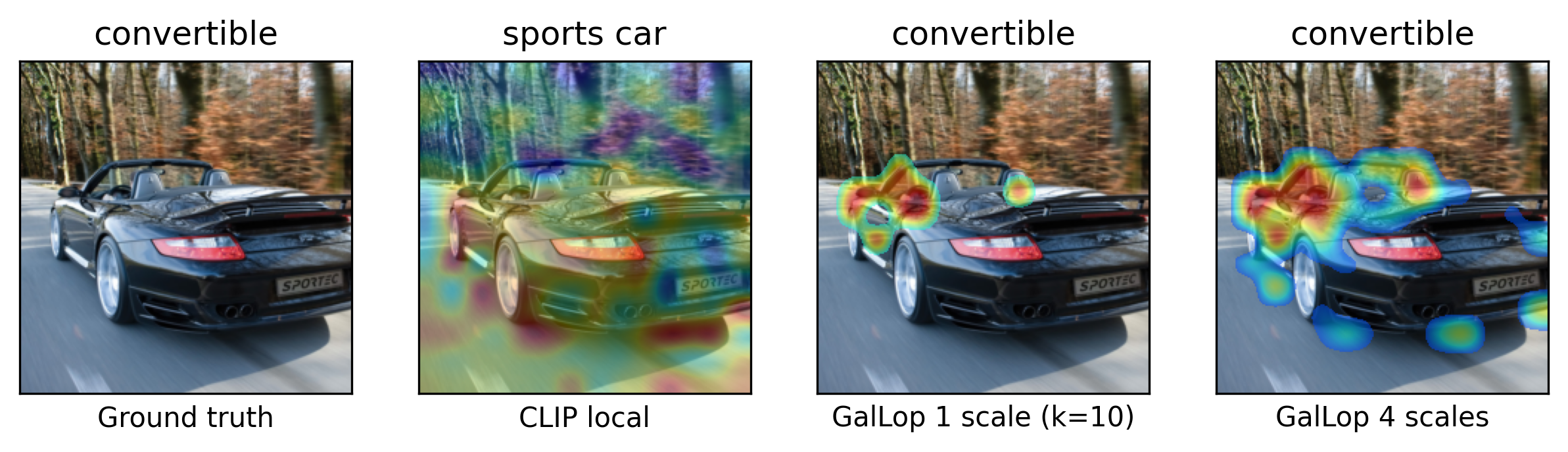} \\
 \includegraphics[width=0.85\linewidth]{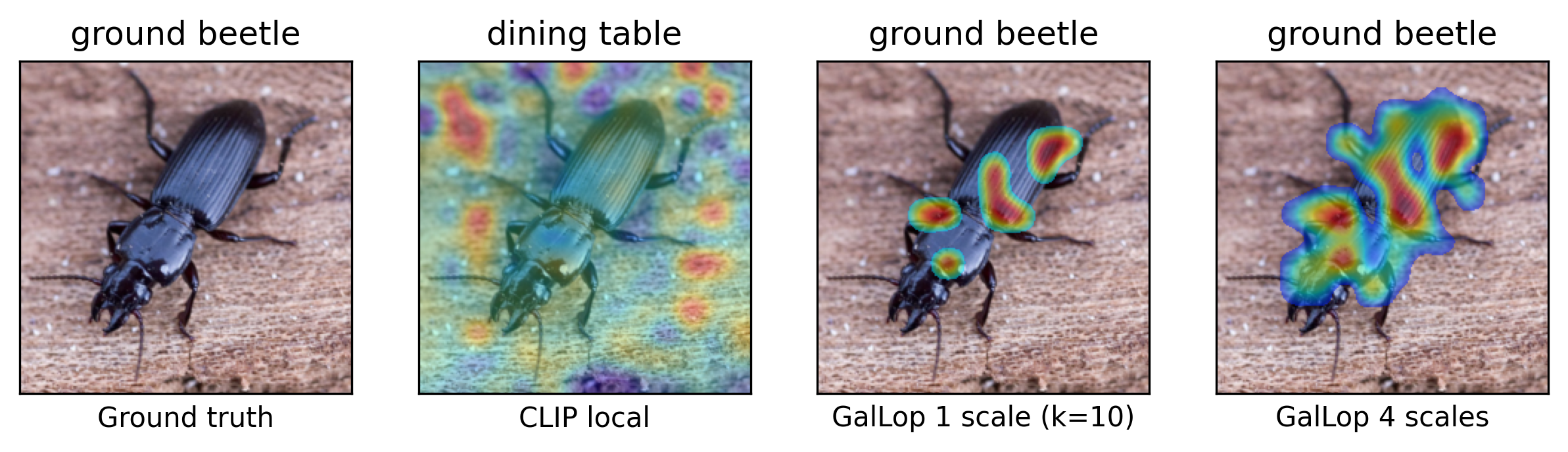} \\
 \includegraphics[width=0.85\linewidth]{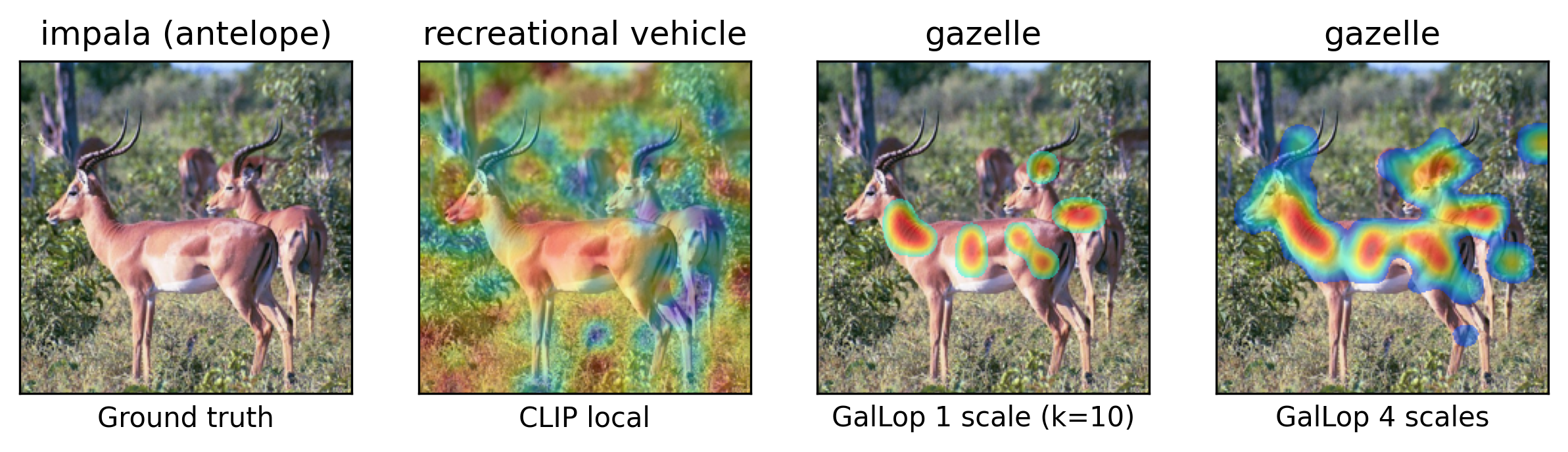} \\
\end{tabular}
\vspace{-1.5em}
\caption{\textbf{Additional qualitative results for \ours.} From left to right, the original image with its ground truth, CLIP local wrong prediction, one scale ($k$=10) of GalLoP\textsubscript{Local} with correct prediction and GalLoP\textsubscript{Local} multiscale.}
\label{fig:supp_qual_results_compa_clip}
\end{figure}

Finally, we display additional qualitative results of GalLoP\textsubscript{Local} on \cref{fig:supp_qual_results_compa_clip}.

\end{document}